%% file: main.tex
\RequirePackage[OT1]{fontenc}
\documentclass[letterpaper, 10 pt, conference]{ieeeconf}

\usepackage{ifthen}
\newcommand\arxivmode{true} 

\IEEEoverridecommandlockouts
\usepackage{graphicx}
\DeclareGraphicsExtensions{.pdf,.png,.jpg}

\usepackage{stfloats}
\let\labelindent\relax
\usepackage[shortlabels]{enumitem}
\usepackage{mwe}
\usepackage{bbm}
\usepackage{amsfonts}
\usepackage{amsmath}
\usepackage{amssymb}
\usepackage{balance}
\usepackage{xcolor}
\usepackage[bookmarks=true]{hyperref}
\usepackage{url}
\usepackage{mathtools}
\usepackage{subfig}
\usepackage[binary-units=true]{siunitx}
\usepackage[ruled,linesnumbered]{algorithm2e}
\usepackage[font=small]{caption}
\usepackage{booktabs}
\usepackage{makecell}
\usepackage{pgfplotstable}
\usepackage{fancyhdr}
\ifthenelse{\equal{\arxivmode}{true}}
{
  \usepackage[frozencache]{minted}
}
{
  \usepackage[finalizecache]{minted}
}

\definecolor{bg}{rgb}{0.95,0.95,0.95}

\makeatletter
\let\NAT@parse\undefined
\makeatother
\usepackage[numbers,sort&compress]{natbib}
\usepackage[capitalise]{cleveref}

\input{content/symbols}

\usepackage[utf8]{inputenc}
\usepackage{pgfplots}
\usepgfplotslibrary{groupplots,dateplot}
\usetikzlibrary{patterns,shapes.arrows}
\pgfplotsset{compat=newest,scaled x ticks=false}

\usepackage{tikz}
\usetikzlibrary{external,positioning}
\usepgfplotslibrary{external}
\newcommand{\edit}[1]{{\color{black}{#1}}}
\newcommand{\blue}[1]{{\color{black}{#1}}}

\title{\bf GIRA: Gaussian Mixture Models for Inference and Robot Autonomy}

\author{Kshitij Goel and Wennie Tabib
\thanks{Project webpage: \url{https://gira3d.github.io/}. The authors are affiliated with the Robotics Institute at Carnegie Mellon University, Pittsburgh, PA 15213, USA.
Emails: \texttt{kshitij@cmu.edu}, \texttt{wtabib@cmu.edu}.%
}
}

\begin{document}
\bstctlcite{IEEEexample:BSTcontrol}
{
\twocolumn[{%
\begin{@twocolumnfalse}
\maketitle
\begin{minipage}{\textwidth}
  \centering
  \ifthenelse{\equal{\arxivmode}{true}}
  {
  \includegraphics[width=0.33\textwidth,height=3cm,trim={0 20 0 0},clip]{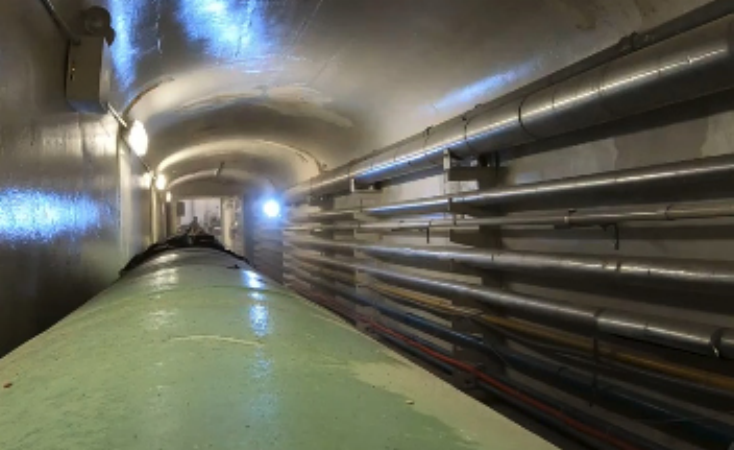}%
  \includegraphics[width=0.33\textwidth,height=3cm,trim={0 20 0 0},clip]{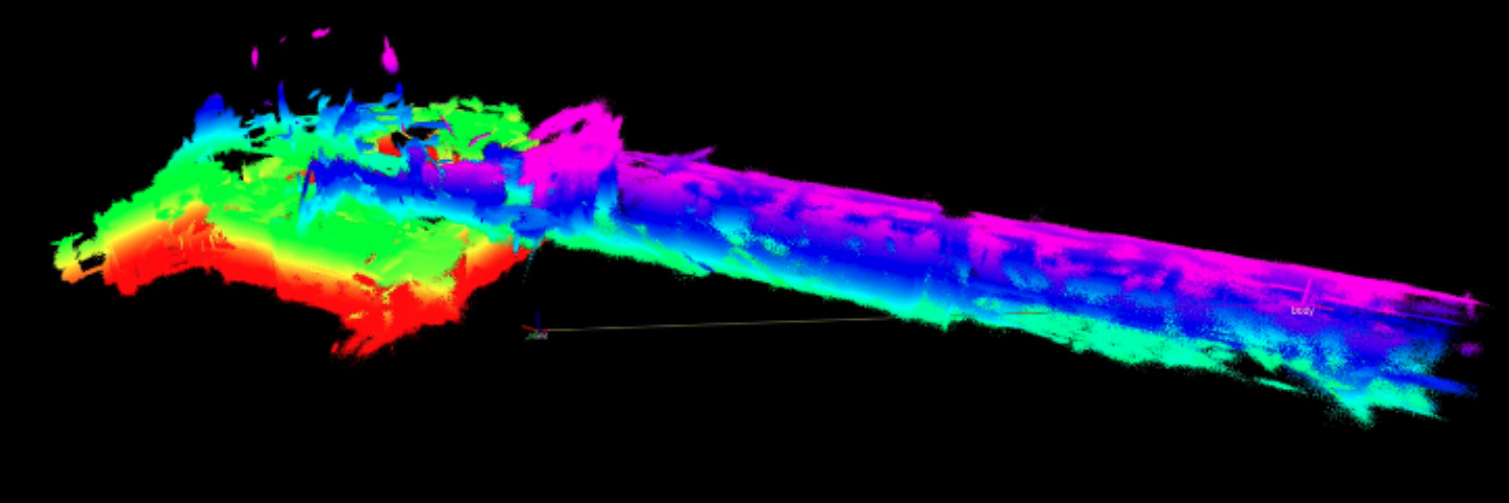}%
  \includegraphics[width=0.33\textwidth,height=3cm,trim={0 20 0 0},clip]{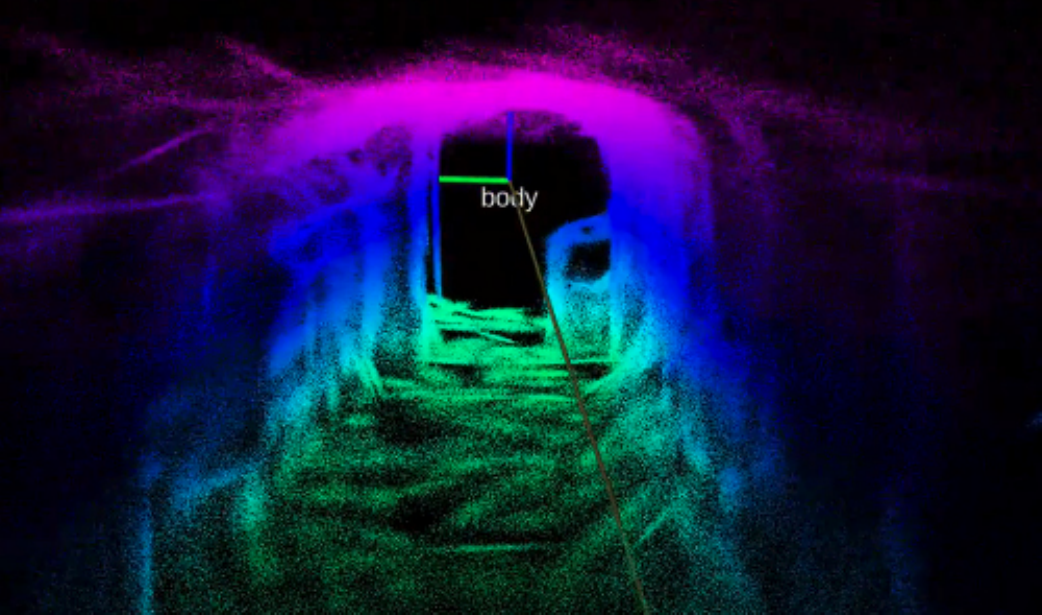}\\
  \includegraphics[width=0.33\textwidth,height=3cm,trim={0 20 0 0},clip]{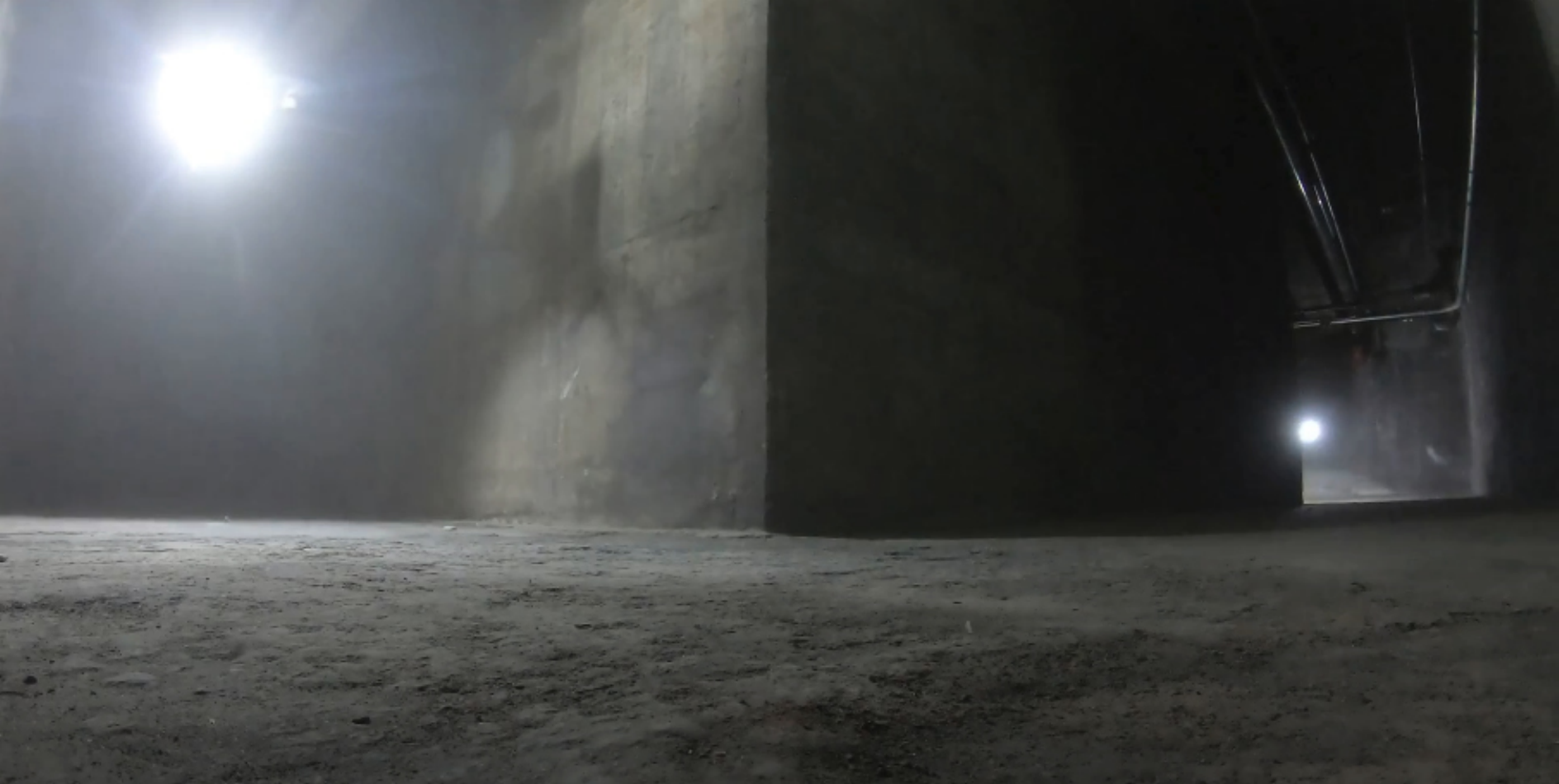}%
  \includegraphics[width=0.33\textwidth,height=3cm,trim={0 20 0 0},clip]{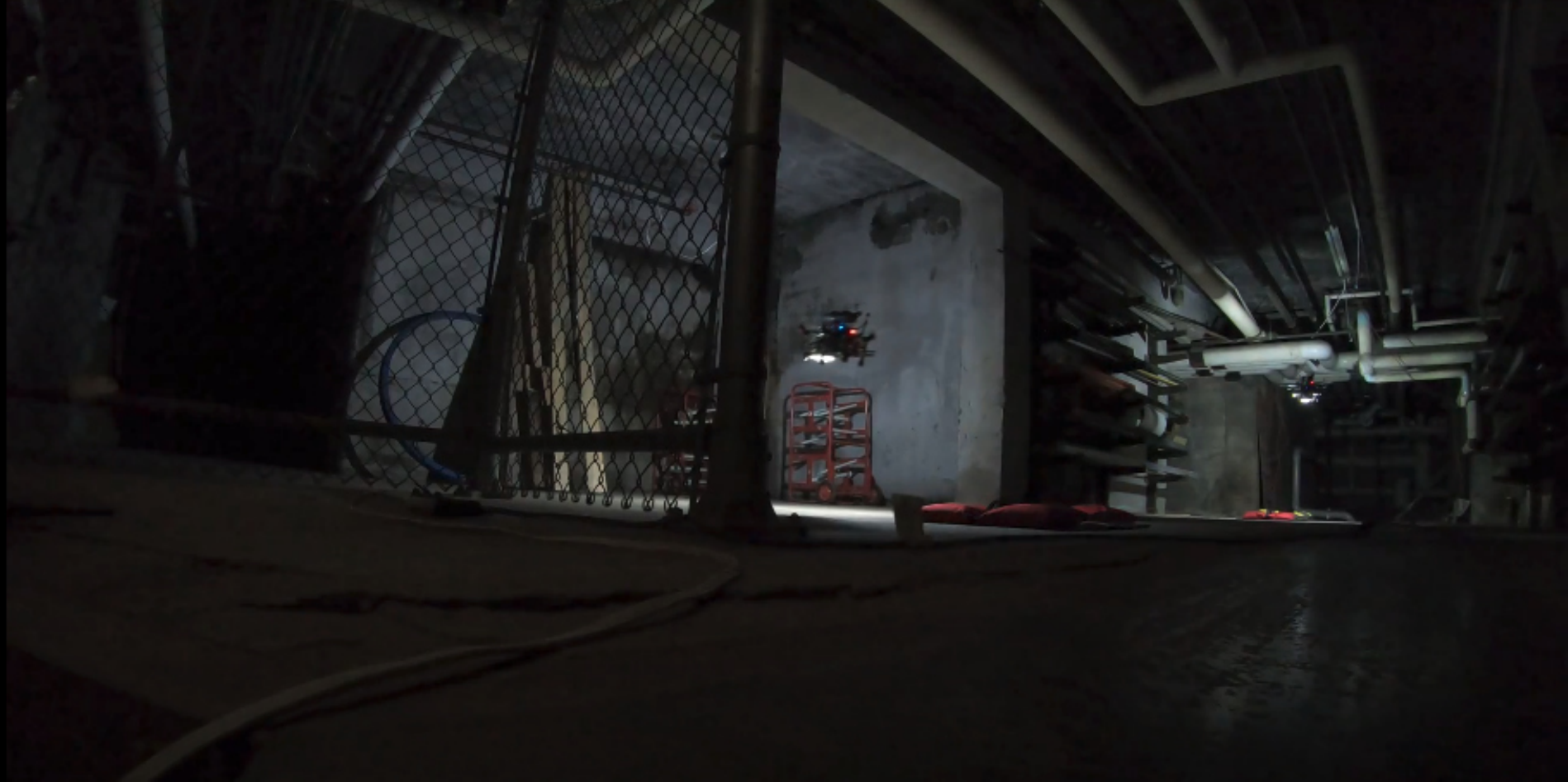}%
  \includegraphics[width=0.33\textwidth,height=3cm,trim={0 20 0 0},clip]{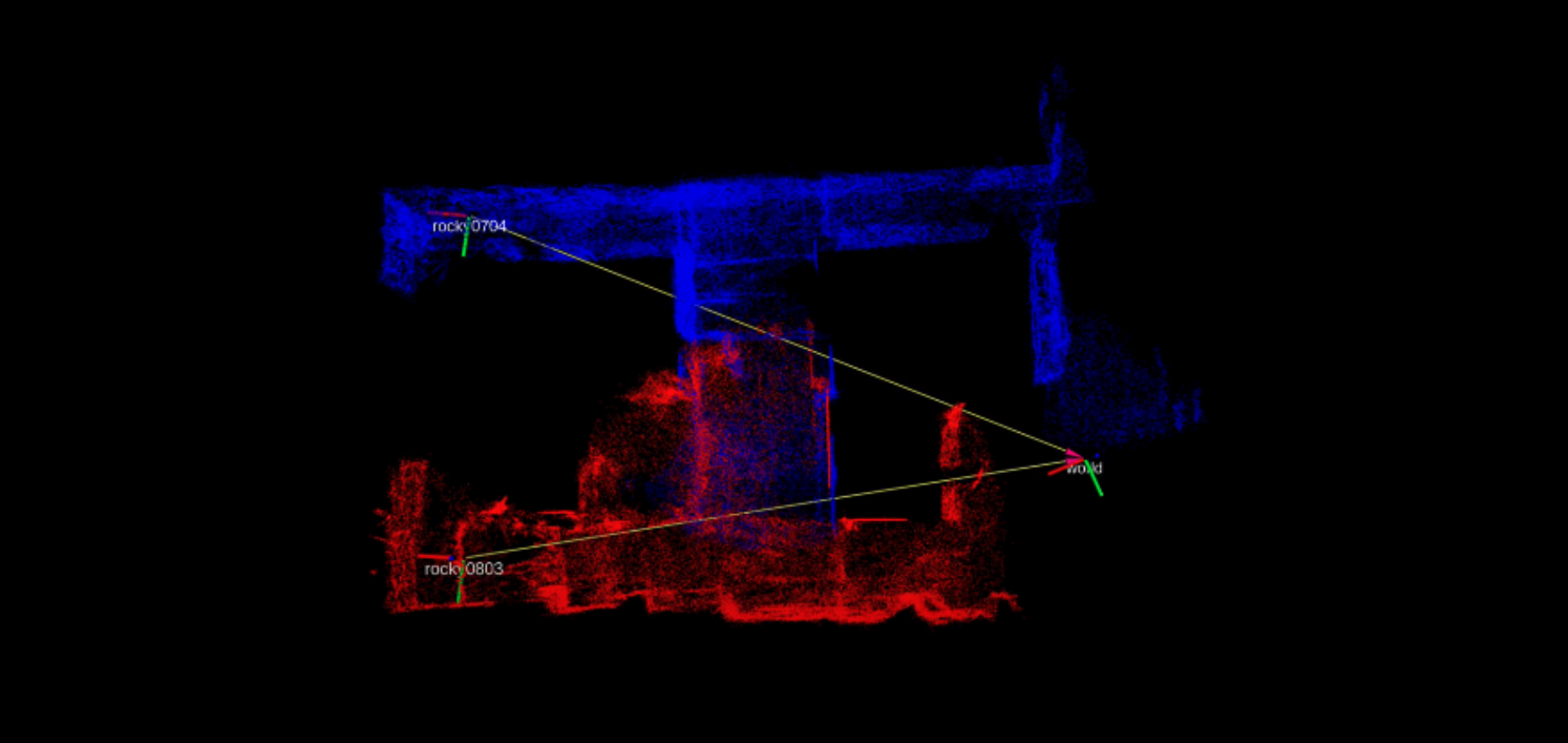}%
  }%
  {
  \includegraphics[width=0.33\textwidth,height=3cm,trim={0 20 0 0},clip]{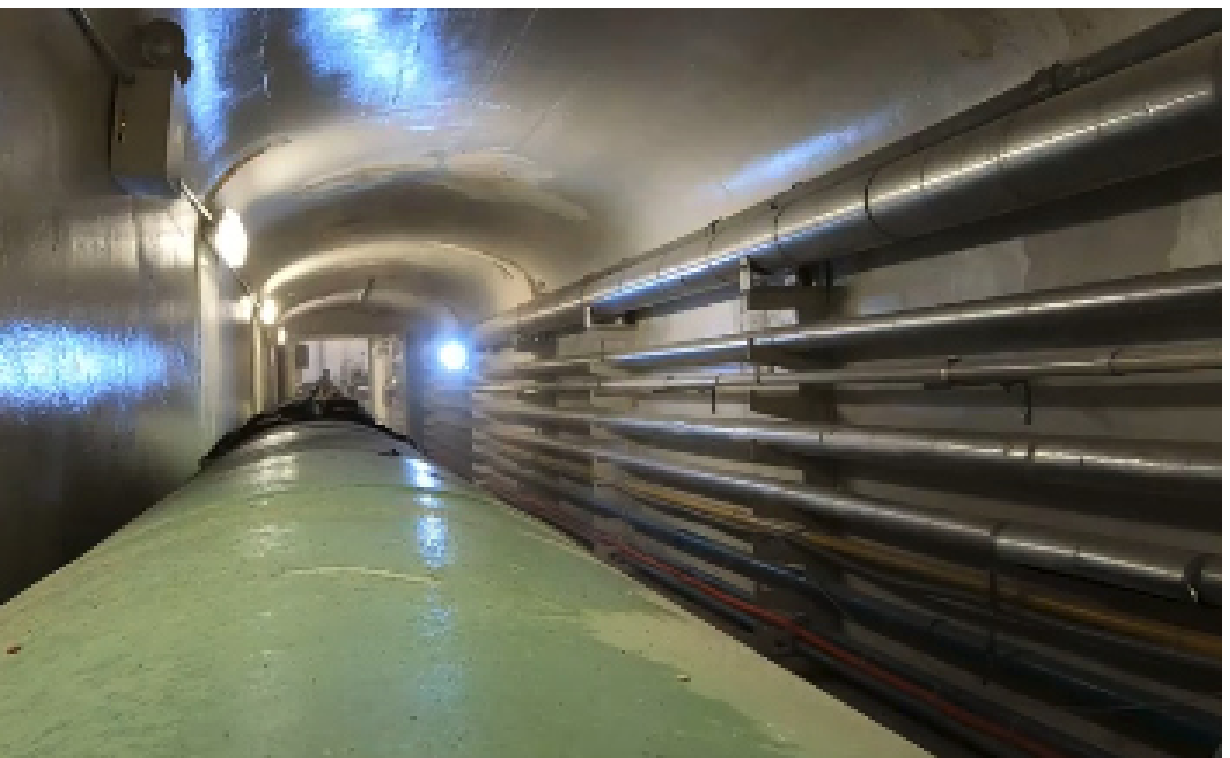}%
  \includegraphics[width=0.33\textwidth,height=3cm,trim={0 20 0 0},clip]{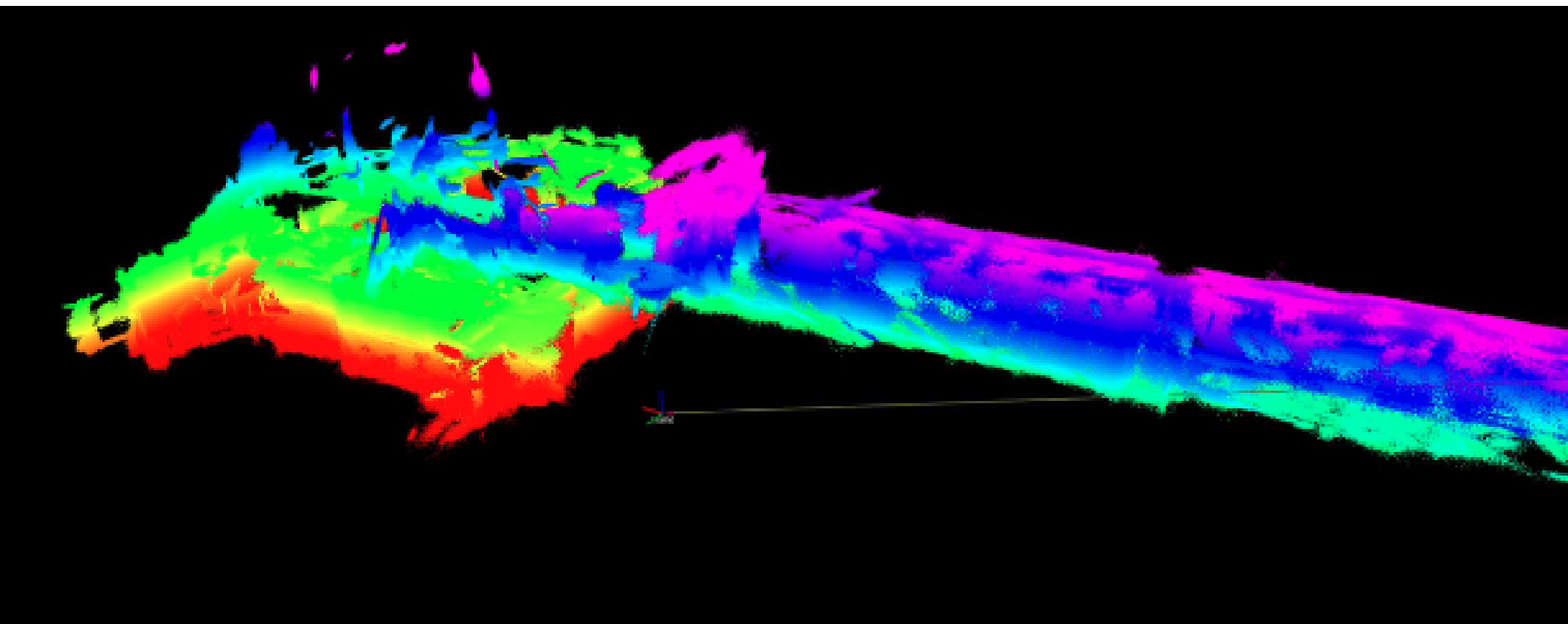}%
  \includegraphics[width=0.33\textwidth,height=3cm,trim={0 20 0 0},clip]{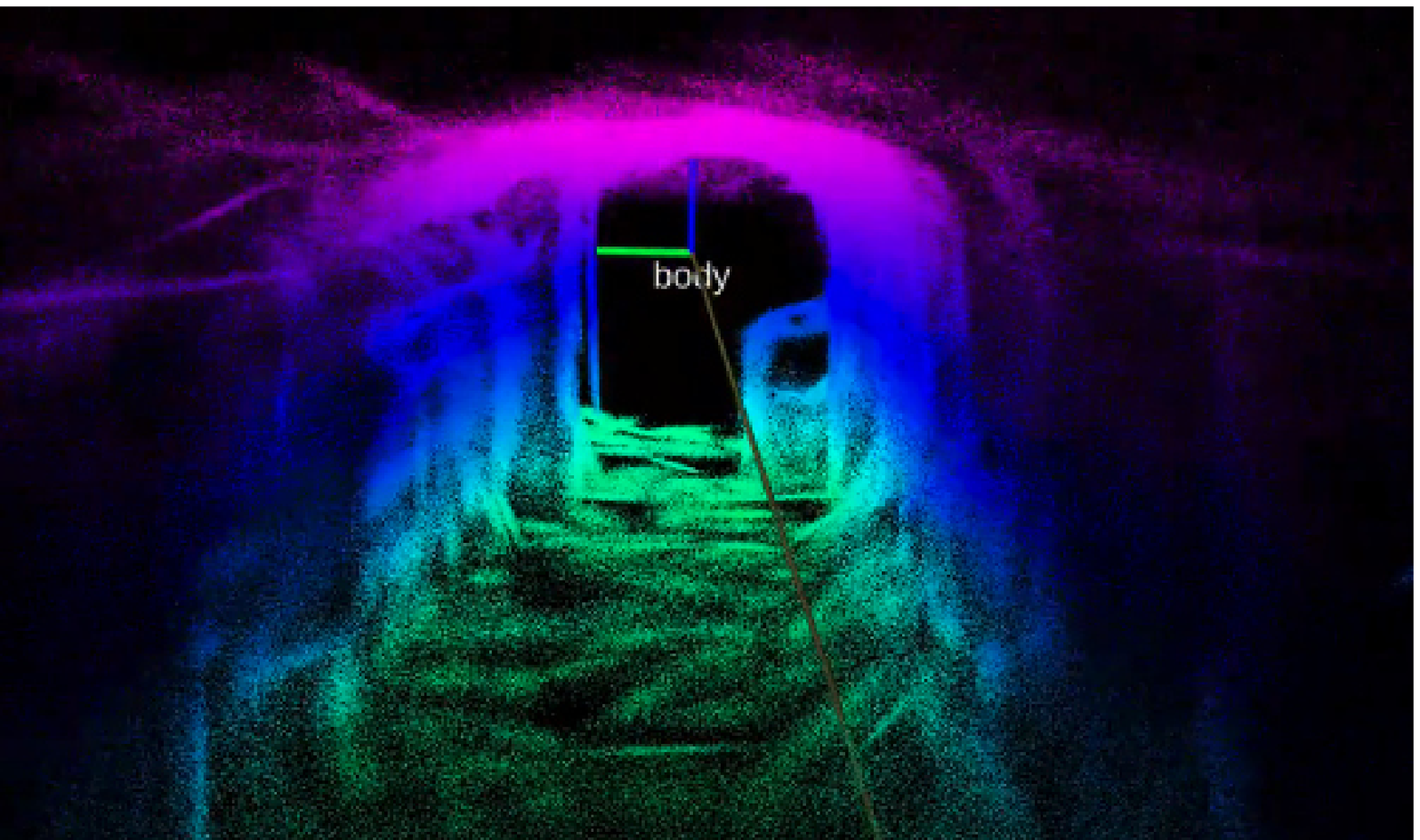}
  \includegraphics[width=0.33\textwidth,height=3cm,trim={0 20 0 0},clip]{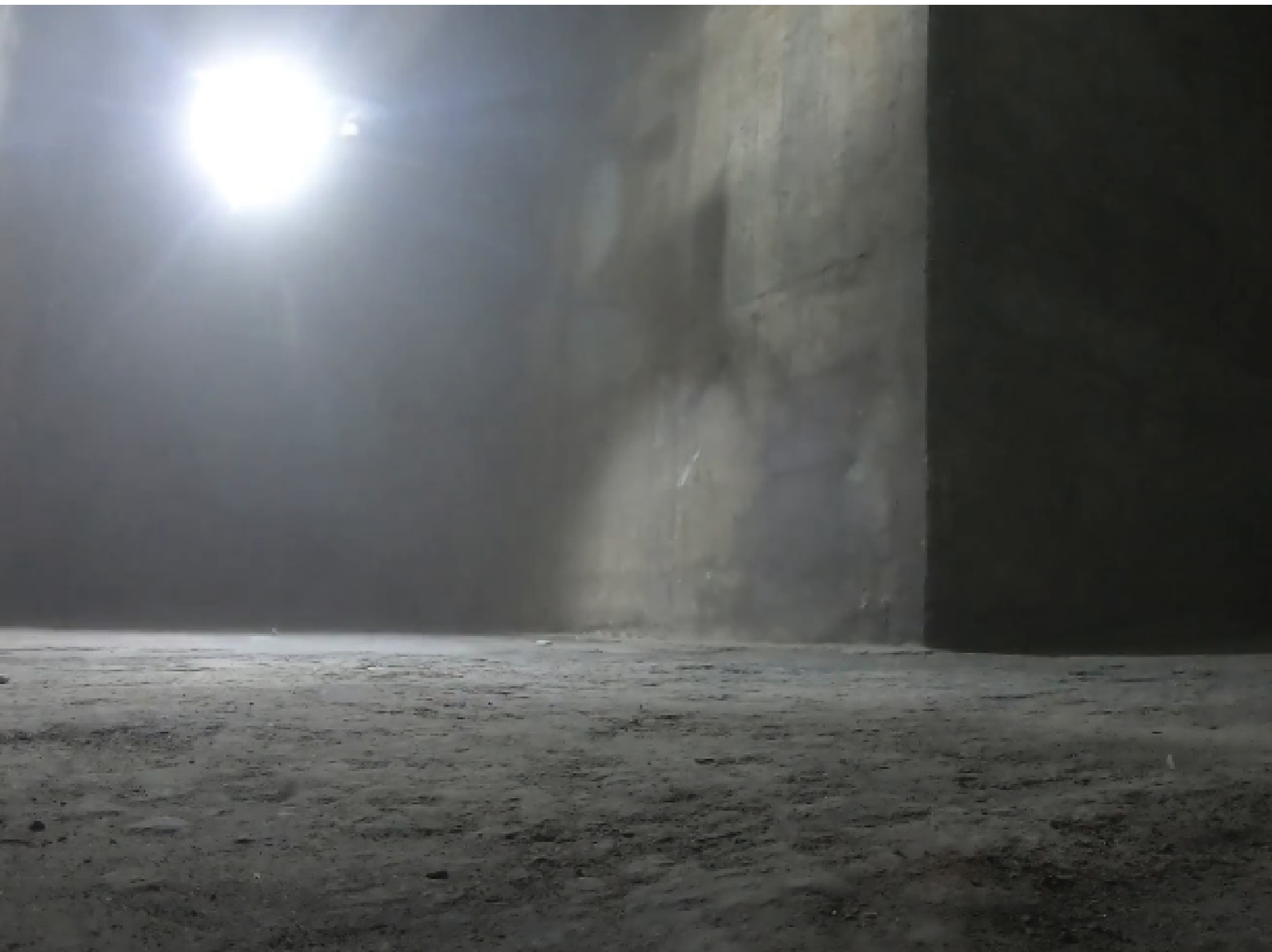}%
  \includegraphics[width=0.33\textwidth,height=3cm,trim={0 20 0 0},clip]{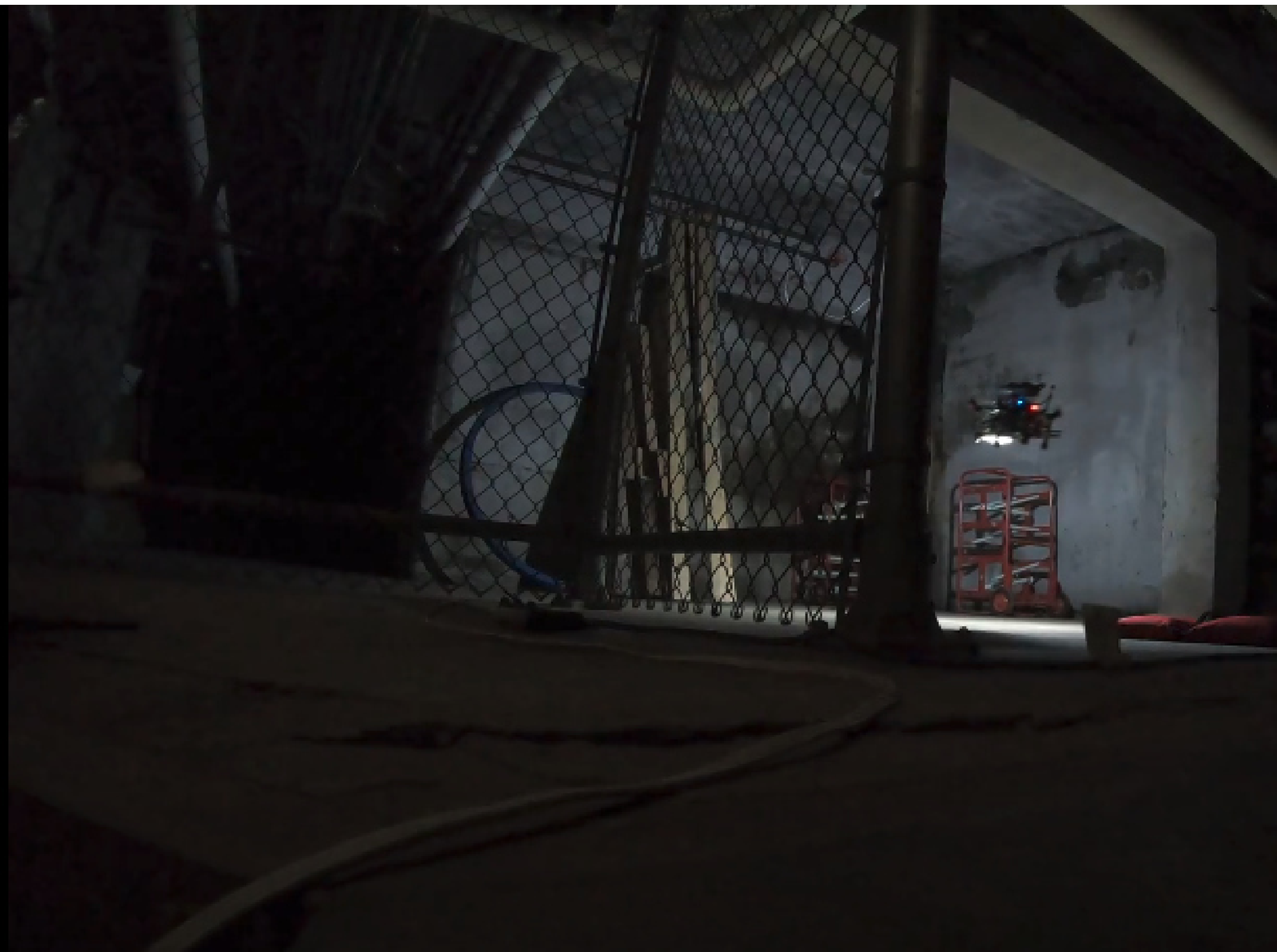}%
  \includegraphics[width=0.33\textwidth,height=3cm,trim={0 20 0 0},clip]{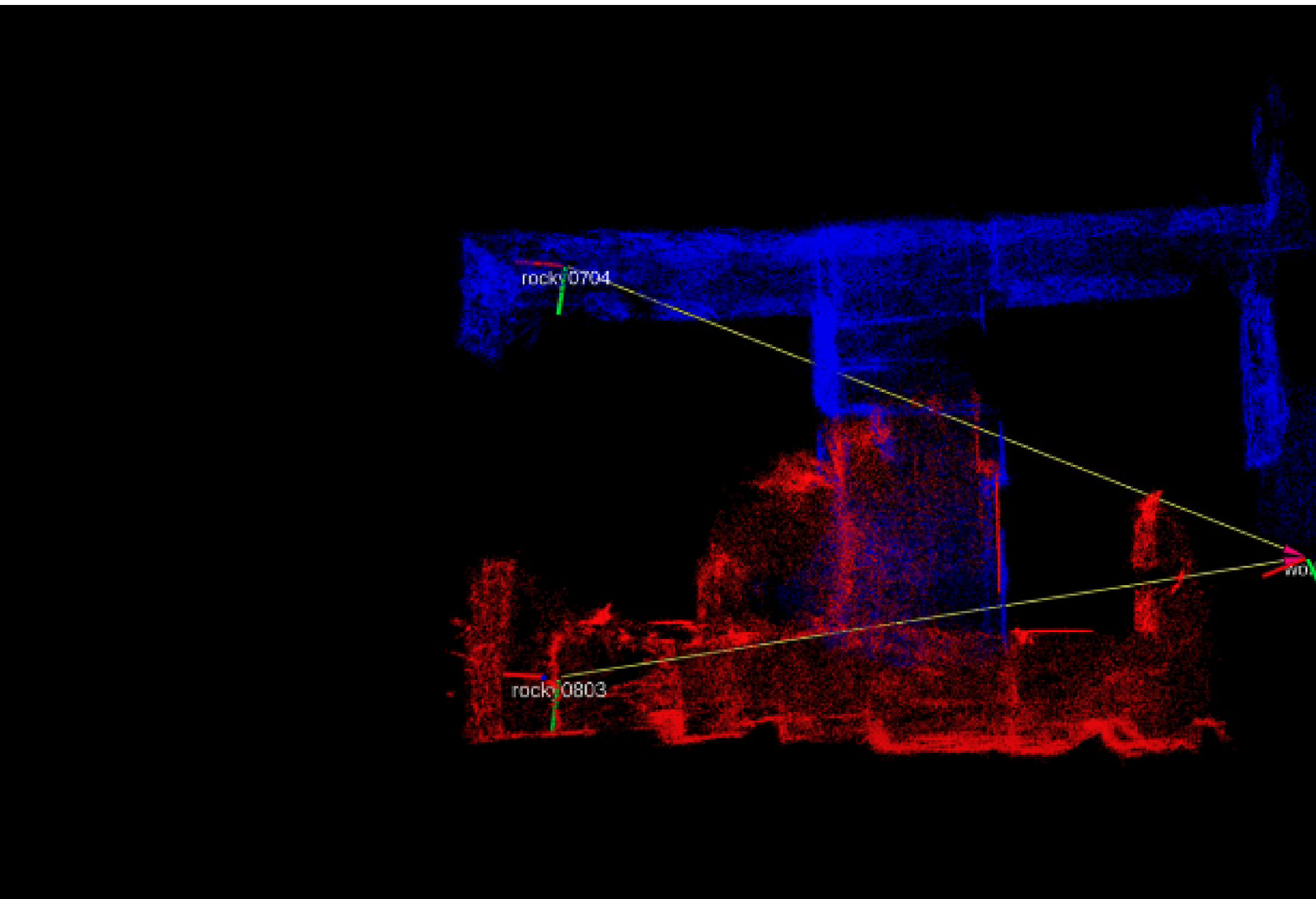}%
  }
  \captionof{figure}{\label{fig:glory-shot} GIRA has been
    deployed on size, weight, and power constrained aerial systems in
    real-world \blue{and} unstructured environments. (Top left) A single
    aerial robot flies through an industrial tunnel and (top center) generates a
    high-fidelity Gaussian mixture model (GMM) map of the environment.
    (Top right) A close-up view of the reconstructed area around the robot.
    (Bottom left and bottom center) A team of two robots fly
    through a dark tunnel environment and produce \blue{a (bottom right)} map\blue{, which}
    is resampled from the underlying GMM and
    colored red or blue according to which robot took the observation.
    Videos of these experiments are available at:
  \url{https://youtu.be/qkbxfxgCoV0} and
  \url{https://youtu.be/t9iYd33oz3g}.}
\end{minipage}
\end{@twocolumnfalse}
}]
{
  \renewcommand{\thefootnote}%
  {\fnsymbol{footnote}}
  \footnotetext[1]{The authors are with The Robotics Institute, Carnegie Mellon University, Pittsburgh, PA 15213 USA
 (email: \{\texttt{kshitij,wtabib}\}\texttt{@cmu.edu}).}
}
}%

\begin{abstract}
\blue{This paper introduces the open-source framework, GIRA, which
implements fundamental robotics algorithms for reconstruction, pose
estimation, and occupancy modeling using compact generative
models. Compactness enables \emph{perception in the large} by ensuring
that the perceptual models can be communicated through low-bandwidth
channels during large-scale mobile robot deployments.  The generative
property enables \emph{perception in the small} by providing
high-resolution reconstruction capability.  These properties address
perception needs for diverse robotic applications, including
multi-robot exploration and dexterous manipulation.}
State\blue{-}of\blue{-}the\blue{-}art perception \blue{systems} construct
\blue{perceptual} models via multiple disparate pipelines that reuse
the same underlying sensor data, which leads to increased computation,
redundancy, and complexity. \blue{GIRA bridges this gap by providing a
unified perceptual modeling framework using Gaussian mixture models
(GMMs) as well as a novel systems contribution, which consists of
GPU-accelerated functions to learn GMMs 10-100x faster compared to
existing CPU implementations. Because few
GMM-based frameworks are open-sourced,
this work seeks to accelerate innovation and broaden adoption of
these techniques.}
\end{abstract}

\section{Introduction}\label{sec:intro}
\input{content/introduction}
\section{Related Work}\label{sec:related-work}
\input{content/related_work}
\section{Design}\label{sec:design}
\input{content/design}
\section{GIRA Framework}\label{sec:approach}
\input{content/approach}
\section{Implementation Details}
\input{content/impl_detail.tex}
\section{Conclusion}
\input{content/conclusion}

{
  \footnotesize
  \bibliographystyle{IEEEtranN}
  \bibliography{refs,do-not-modify}
}

\end{document}

%% file: content/symbols.tex
\newcommand{\datasetY}{\mathcal{Y}}
\newcommand{\pointx}{\mathbf{x}}

\newcommand{\weight}{\pi}

\newcommand{\mean}{\boldsymbol{\mu}}

\newcommand{\cov}{\mathbf{\Sigma}}

\newcommand{\gaussian}[4][\mathcal{N}]{\ensuremath{ {#1} ({#2} \mid {#3}, {#4})}}

\newcommand{\findexset}{\mathcal{B}}

\newcommand{\nComponents}{|\findexset|}

\newcommand{\resp}{\gamma}

\newcommand{\lowerL}{\text{L}}
\newcommand{\precchol}{\text{P}}

\definecolor{darkgray176}{RGB}{176,176,176}
\definecolor{green}{RGB}{0,128,0}
\definecolor{lightgray204}{RGB}{204,204,204}


%% file: content/introduction.tex
\blue{To navigate in and interact with the world, robots acquire,
assimilate, and respond to sensor data. Models that enable perception
in the large and small~\cite{Bajcsy:2018tf} are amenable to diverse
robotics applications and have the potential to drastically increase
robotic capabilities while addressing limitations in the way complex
perception systems are developed today.}

Recent large-scale robotic exploration deployments, like the DARPA
Subterranean (Sub-T) Challenge~\cite{chung2022into}, have highlighted
the need for map compression \blue{to} increase the \blue{exploration
rate, an example of perception in the large,} by facilitating information sharing. Further, state-of-the-art
perception systems typically leverage separate concurrent perceptual
processing pipelines, which increases computation, redundancy, and
complexity~\cite{Eckart-2017-104773}. For example, the highly
sophisticated perception module of the NeBula system
architecture~\citep{agha_nebula_2022-1} processes the same LiDAR data
repeatedly (e.g., odometry, SLAM, terrain mapping, etc.), which is
inefficient. Instead, what is needed is a
unified framework for common perceptual processing elements, which is
compact, generative, and amenable for deployment on low-power embedded
systems~\cite{Eckart-2017-104773}.

Gaussian mixture models (GMMs) provide high-fidelity and
communication-efficient point cloud modeling and
inference~\citep{corah_communication-efficient_2019} in real-world
environments~\citep{tabib_autonomous_2021}. \blue{Recent works have demonstrated
precise, high-fidelity representation of fine details required
for perception in the small~\cite{goel_probabilistic_2023}}. However, there are few
open-source implementations, which poses a barrier to broad adoption
by the general robotics community. To bridge this gap, this paper
introduces GIRA, an open-source, \blue{unified} framework
\blue{(\cref{fig:glory-shot})} for \blue{point cloud modeling,
occupancy modeling, and pose estimation using GMMs based
on}~\citep{tabib_-manifold_2018,tabib_simultaneous_2021,tabib_autonomous_2021}.
\blue{In addition, GIRA includes a novel systems contribution, which
consists of} GPU-accelerated functions to learn GMMs 10-100x faster
compared to \blue{existing} CPU implementations. The software and
associated datasets are open-sourced\footnote{Project webpage:
\url{https://github.com/gira3d}} to accelerate innovation and adoption of these techniques.

%% file: content/related_work.tex
This section reviews open-source perception frameworks for
compact, high-resolution point cloud modeling, pose estimation, and occupancy
modeling for robotics applications. These works are compared and contrasted with
GIRA.



The Normal Distributions Transform
  (NDT)\footnote{\url{https://github.com/OrebroUniversity/perception\_oru}}
framework was introduced by~\citet{biber_normal_2003} for scan registration
and later extended to 3D registration~\citep{magnusson_scan_2007} and occupancy modeling~\citep{saarinen_normal_2013}.
~\citet{goel_probabilistic_2023} demonstrate
that NDTMap provides higher representation fidelity compared to
Octomap\footnote{\url{https://github.com/Octomap/octomap}}~\citep{hornung_octomap_2013},
but at the cost of increased disk storage requirements.
While NDTMap
provides distribution to distribution registration~\cite{stoyanov2012fast},
Octomap does not provide analogous functionality. In contrast to these representations, GIRA provides higher
memory-efficiency and surface reconstruction
fidelity~\citep{goel_probabilistic_2023} as well as
distribution to distribution
registration~\citep{tabib_-manifold_2018,tabib_simultaneous_2021}.  Further,
NDTMap provides a CPU implementation, while GIRA provides both CPU and GPU
implementations for multimodal environment modeling.

\citet{oleynikova_voxblox_2017} develop Voxblox, which uses Truncated
Signed Distance Fields (TSDFs), for high-resolution
reconstruction and occupancy mapping. The weights for the
TSDFs are stored in a coarse fixed-resolution regular
grid. Voxblox grows dynamically, but suffers from the
same memory-efficiency limitation as the NDTMap. In contrast, the GIRA
framework enables high-resolution surface reconstruction without a
pre-specified size or a fixed-resolution discretization of the point
cloud model.  Like GIRA, Voxblox provides
CPU\footnote{\url{https://github.com/ethz-asl/voxblox}} and
GPU\footnote{\url{https://github.com/nvidia-isaac/nvblox}}
implementations as well as a method to localize within the
representation using submaps~\cite{reijgwart_voxgraph_2020}.

\citet{duberg_ufomap_2020} propose UFOMap, which improves upon
Octomap by providing an explicit representation of unknown
space and introduces Morton codes for faster tree traversal. An
open-source CPU implementation of
UFOMap\footnote{\url{https://github.com/UnknownFreeOccupied/ufomap}}
is available; however, the implementation does not provide
functionality to localize within the map, like GIRA, Voxgraph, or NDTMap.

\citet{vespa_efficient_2018} introduce the Supereight mapping
framework, which consists of two dense mapping methods a
TSDF-based implicit map and an explicit spatial occupancy
map. In follow-on work~\citep{funk_multi-resolution_2021}, which
leverages multi-resolution grids, the authors demonstrate that the
TSDF-based method yields superior reconstruction compared to
UFOMap. Supereight enables frame-to-model
point cloud registration via Iterative-Closest-Point (ICP) alignment.
However, Supereight uses RAM to assess memory efficiency,
but does not provide statistics on space required to store the
representation to disk. The CPU implementation is available for
Supereight\footnote{\url{https://bitbucket.org/smartroboticslab/supereight2}}
as open-source software.

~\citet{reijgwart_efficient_2023} have recently proposed the Wavemap
hierarchical volumetric representation, which uses wavelet compression for higher memory savings
compared to Voxblox, Supereight, and Octomap. Like other discrete mapping
methods, the highest resolution of the hierarchical map is set by
the user and fixed during robot operation. In contrast, the SOGMM method in GIRA provides
the ability to adapt the fidelity
of the model according to the complexity of the scene, without utilizing a hierarchical approach~\citep{goel_probabilistic_2023}.
The CPU implementation for Wavemap is available as open-source software\footnote{\url{https://github.com/ethz-asl/wavemap}}
and includes some additional improvements (e.g., the use of OpenVDB~\citep{museth_openvdb_2013}
instead of the octree data structure used in the original paper).
The approach, however, does not provide a method to estimate pose.

\citet{doherty_bayesian_2017} propose BGKOctomap, an extension
to Octomap, which utilizes nonparametric Bayesian kernel
inference for continuous-space occupancy modeling. The method
is improved by modeling sensor rays as continuous free-space
observations~\citep{doherty_learning-aided_2019}. The
CPU\footnote{\url{https://github.com/RobustFieldAutonomyLab/la3dm}}
implementations of these variants are available as open-source
software.

\citet{eckart_accelerated_2016} present compact modeling of point
clouds using GMMs to estimate pose.
The approach is extended to a hierarchical formulation
to improve memory-efficiency and
accuracy~\citep{eckart_hgmr_2018}.~\citet{dhawale_efficient_2020} present an alternative
hierarchical approach, which equally weights the Guassian distributions,
for high-resolution surface mapping.~\citet{srivastava_efficient_2019} present another
hierarchical approach but utilize the Expectation-Maximization
algorithm to assign non-uniform weights depending on the local density
of point cloud data. Common to these approaches is the need to set
model complexity criteria such as image patch
size~\citep{dhawale_efficient_2020}, component splitting
threshold~\citep{eckart_hgmr_2018}, and model fidelity
threshold~\citep{srivastava_efficient_2019}. Further, to the best of
our knowledge, there are no open-source software implementations of
existing GMM-based point cloud modeling methods. In contrast, we
provide open-source CPU and GPU implementations of our work
in~\citep{goel_probabilistic_2023}, which leverages
information-theoretic learning to
adjust the model complexity. Finally, to demonstrate the occupancy
modeling and pose estimation capabilities using GMM-based models, we
provide CPU implementations based
on~\citep{tabib_-manifold_2018,tabib_simultaneous_2021,tabib_autonomous_2021}.

%% file: content/design.tex
We envision the GIRA framework to be used in robotics research where
3D perception tasks must be executed in real-time and algorithms for
these tasks should be easy to prototype.

Popular robotics software packages like
Bullet~\citep{coumans2021}, Drake~\citep{drake}, and
TensorFlow~\citep{abadi2016tensorflow} provide low-level programming language
support for high-performance, real-time operation and high-level programming
language bindings for ease of prototyping.  We follow the same model with GIRA
where key algorithms are implemented in C/C++/CUDA with Python bindings. For
C/C++, we use the C/C++17 standard. For GPU support, CUDA version 10.4 and above
is required.


To enable message passing between different software
systems, most robotics applications use the Robot Operating System
(ROS)~\citep{Quigley09} and its successor ROS2~\citep{macenskiros2}. The GIRA
framework is structured using Collective Construction (\texttt{colcon}) packages
to help \edit{the} robotics community easily integrate GIRA within their ROS and ROS2
workspaces. For low-level code and bindings, GIRA utilizes CMake for
compilation support on both Linux and macOS. Python virtual environments are
used to isolate executables.

For 3D perception tasks, visualization is an important
capability for debugging research code. GIRA provides interfaces to the
Open3D~\citep{zhou_open3d_2018} visualization tools for this purpose.
Furthermore, developers can leverage tools like RViz2 from ROS2 after integrating
\texttt{colcon} packages from GIRA.

%% file: content/approach.tex
The GIRA framework consists of three components: (1)
\href{https://github.com/gira3d/gira3d-reconstruction}{GIRA Reconstruction},
(2) \href{https://github.com/gira3d/gira3d-registration}{GIRA Registration},
and (3) \href{https://github.com/gira3d/gira3d-occupancy-modeling}{GIRA
Occupancy Modeling}. This section provides an overview of these three
components.

\subsection{GIRA Reconstruction\label{ssec:gira3d-reconstruction}}
\input{content/gira3d_reconstruction}

\input{content/gira3d_registration}

\subsection{GIRA Occupancy Modeling\label{ssec:gira3d-occupancy}}
\input{content/gira3d_occupancy_modeling}

%% file: content/gira3d_reconstruction.tex
\begin{figure*}[t]
      \centering
      \ifthenelse{\equal{\arxivmode}{true}}
      {
      \subfloat[Input\label{sfig:input}]{\includegraphics[width=0.33\textwidth,trim=200 20 230 150,clip]{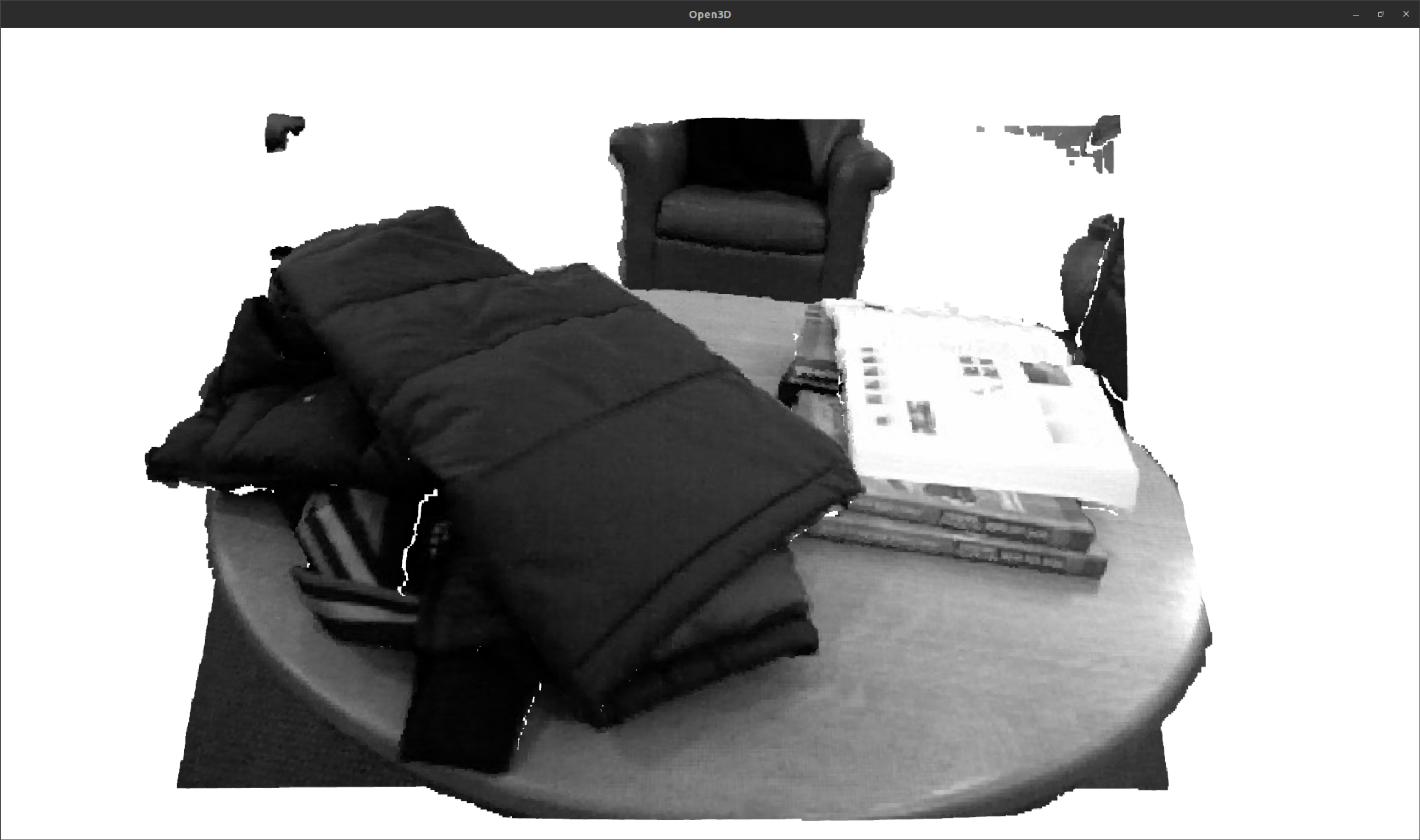}}%
      \subfloat[Resampled\label{sfig:resampled}]{\includegraphics[width=0.33\textwidth,trim=200 20 230 150,clip]{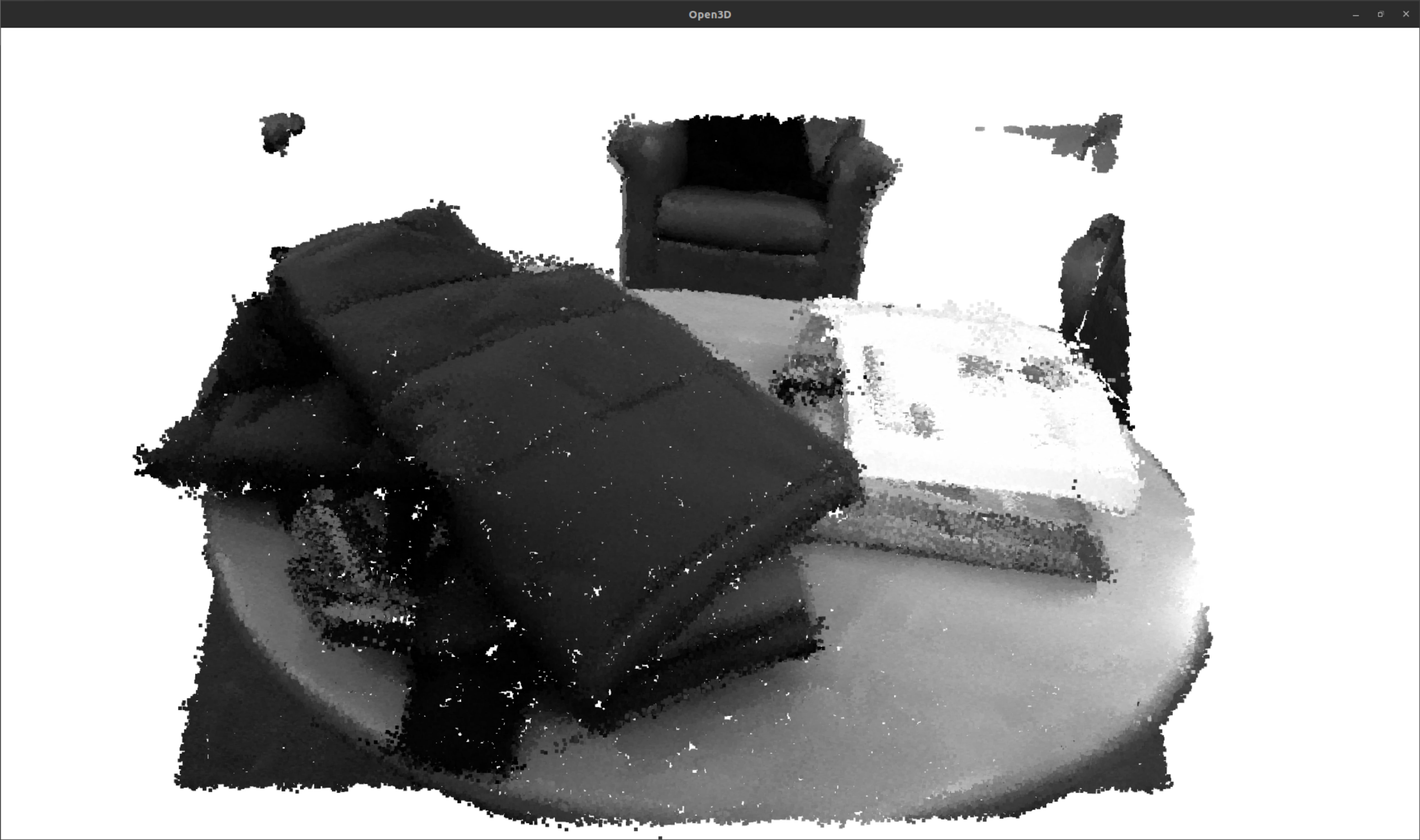}}%
      \subfloat[Intensity Inference\label{sfig:inferred}]{\includegraphics[width=0.33\textwidth,trim=200 20 230 150,clip]{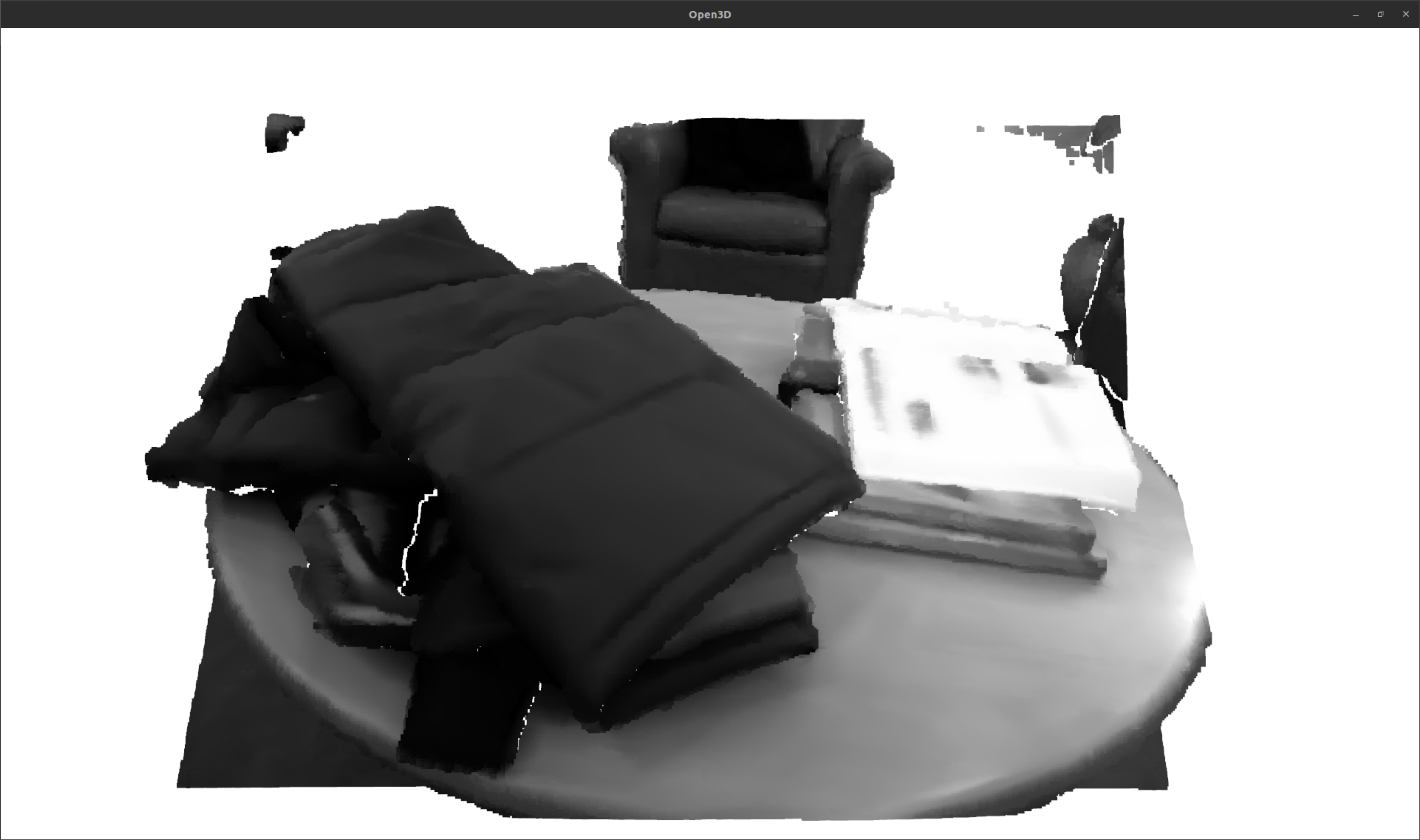}}
      }
      {
      \subfloat[Input\label{sfig:input}]{\includegraphics[width=0.33\textwidth,trim=200 20 230 150,clip]{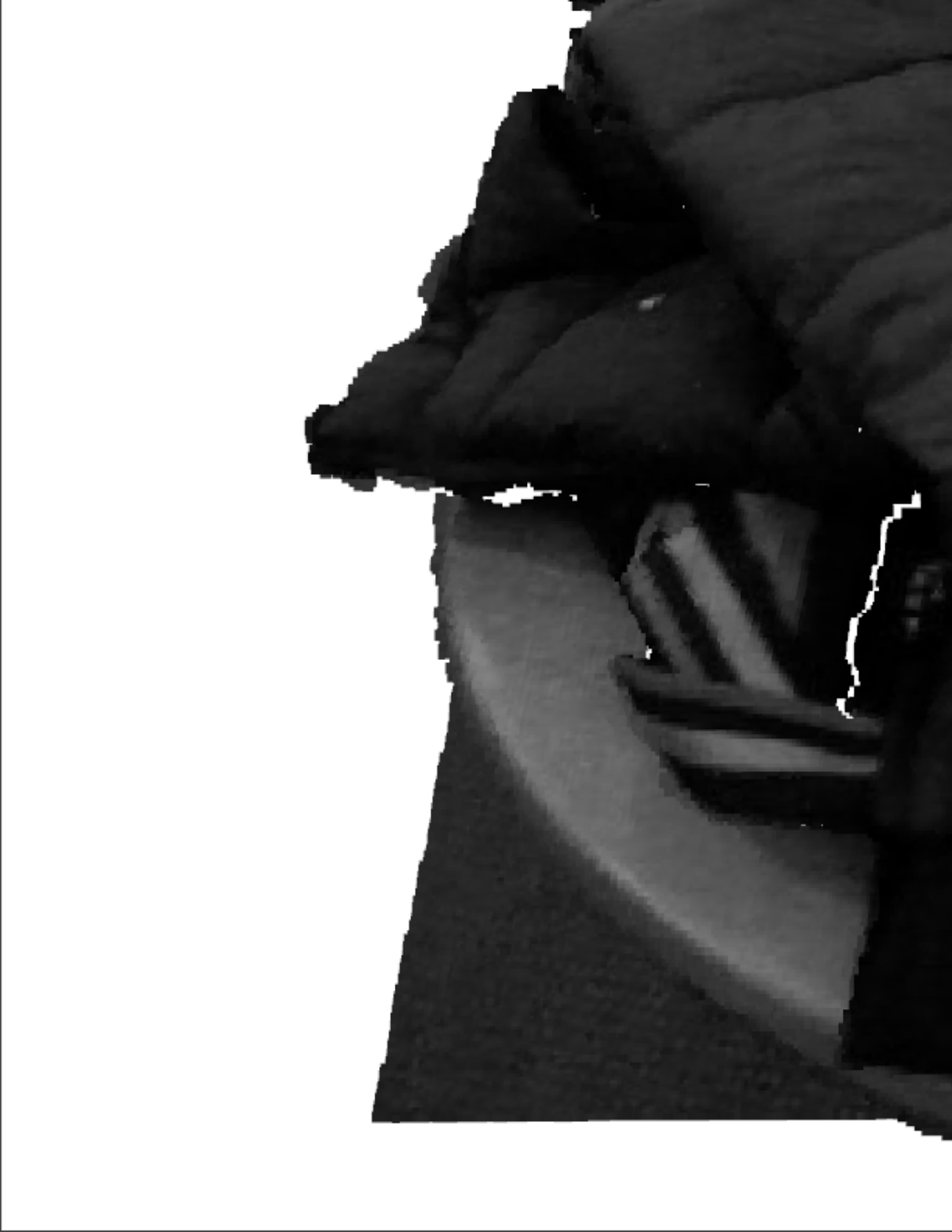}}%
      \subfloat[Resampled\label{sfig:resampled}]{\includegraphics[width=0.33\textwidth,trim=200 20 230 150,clip]{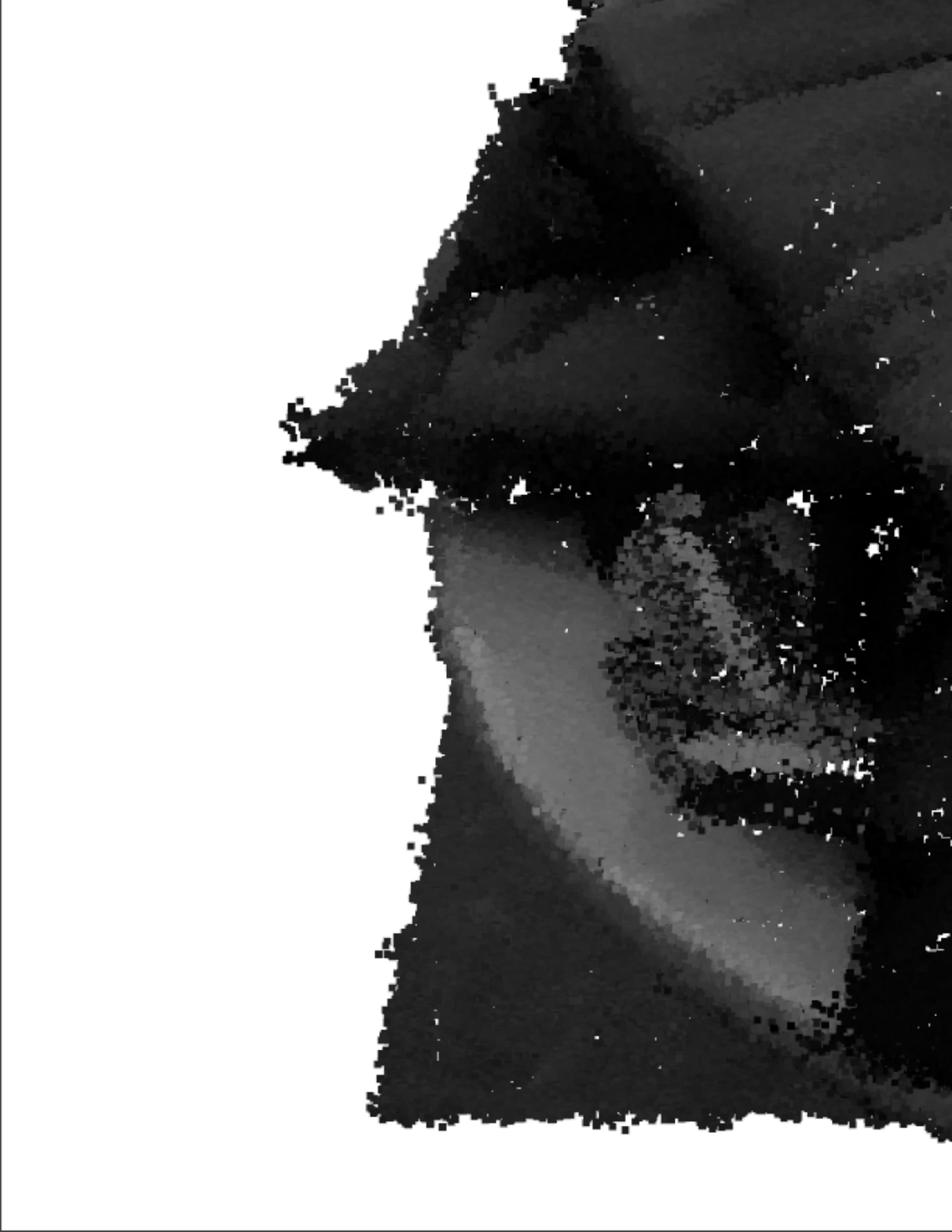}}%
      \subfloat[Intensity Inference\label{sfig:inferred}]{\includegraphics[width=0.33\textwidth,trim=200 20 230 150,clip]{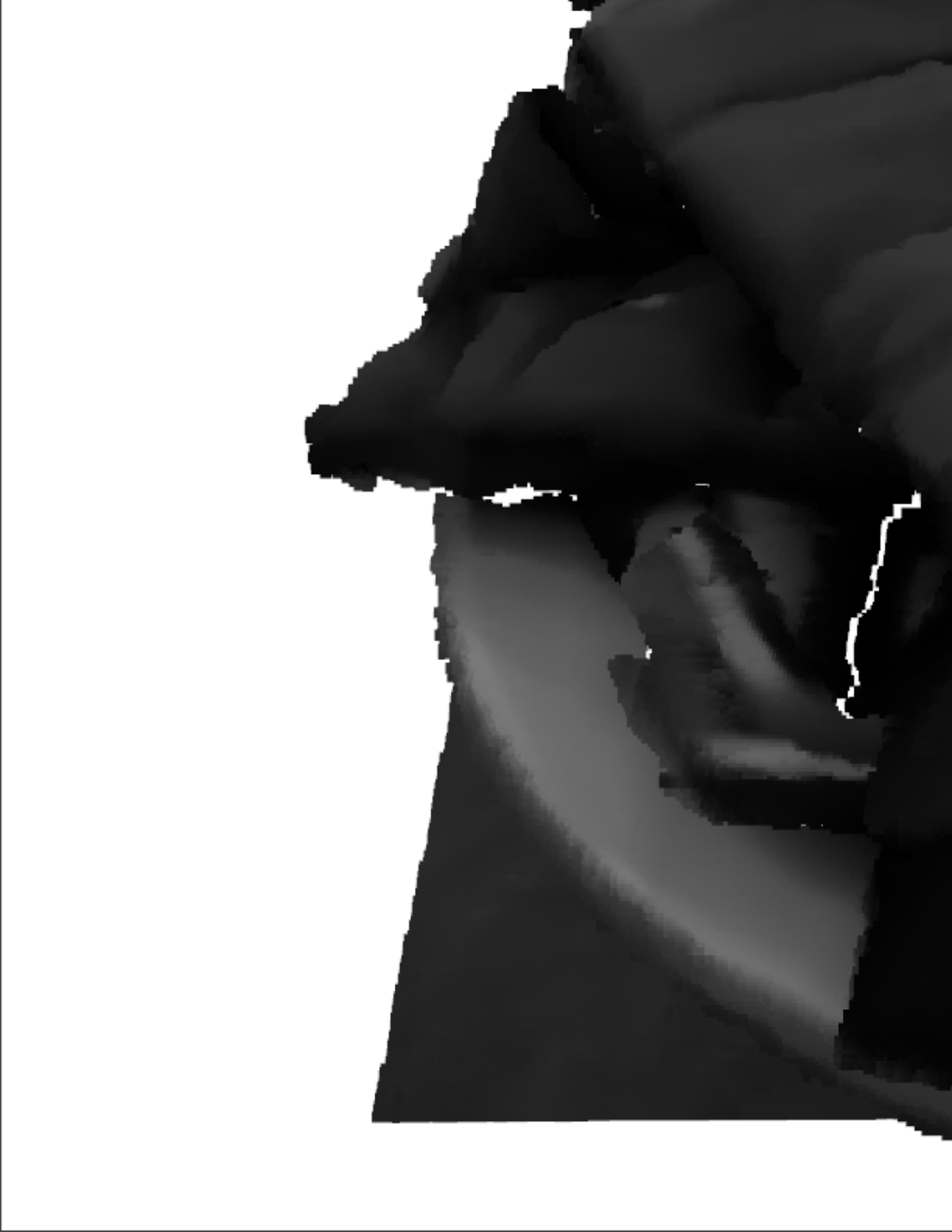}}
      }
      \caption{\label{fig:gira3d-recon}An example workflow for GIRA
            Reconstruction~\cref{ssec:gira3d-reconstruction}. The input is a
            depth-intensity point cloud shown in~\protect\subref{sfig:input}. The
            resulting model can be resampled to generate novel 4D
            points~\protect\subref{sfig:resampled} or be used to infer expected intensity
            values at known 3D locations~\protect\subref{sfig:inferred}.}
\vspace{-0.5cm}
\end{figure*}

Given time-synchronized depth and intensity images with known
pose estimates, GIRA Reconstruction creates a Self-Organizing Gaussian Mixture Model
(SOGMM)~\citep{goel_probabilistic_2023} that is:
\begin{enumerate}
      \item \textbf{Continuous}, the point cloud is represented with a 4D GMM which
            is a linear combination of continuous functions (Gaussian distributions).
      \item \textbf{Probabilistic}, the 4D GMM captures the variance and
            expected values for point locations and intensity values.
      \item \textbf{Generative}, the 4D GMM enables fast sampling of point locations
            and intensity values from the model.
      \item \textbf{Compact}, since the number of parameters required to represent
            the 4D GMM is much lower compared to the point cloud itself.
      \item \textbf{Adaptive}, the number of Gaussian distributions within the 4D
            GMM are automatically estimated from the scene complexity of the
            underlying sensor data.
\end{enumerate}

GIRA Reconstruction utilizes
\href{http://www.open3d.org/}{Open3D}~\citep{zhou_open3d_2018} for point cloud
loading, writing, and visualization. We use
\href{https://numpy.org/}{NumPy}~\citep{harris_array_2020} for interfacing with
\href{http://eigen.tuxfamily.org/}{Eigen}~\citep{eigenweb} via
\href{https://pybind11.readthedocs.io/en/stable/index.html}{Pybind11}~\citep{pybind11}.

GIRA Reconstruction contains CPU and GPU
implementations for SOGMM\footnote{Detailed tutorials are available at
      \url{https://gira3d.github.io/docs/index.html}.}.  Both implementations can be
accessed via a Python interface:
\begin{minted}[bgcolor=bg,breaklines,fontsize=\footnotesize]{python}
from sogmm_py.sogmm import SOGMM

# SOGMM of pointcloud on CPU
sg_cpu = SOGMM(bandwidth=0.015, compute='CPU')
mcpu = sg_cpu.fit(pointcloud)

# SOGMM of pointcloud on GPU
sg_gpu = SOGMM(bandwidth=0.015, compute='GPU')
mgpu = sg_gpu.fit(pointcloud)
\end{minted}
where, \texttt{pointcloud} is a NumPy array and \texttt{bandwidth}
is the bandwidth of the kernel used for the Gaussian Blurring Mean Shift (GBMS)
within the SOGMM algorithm~\citep{goel_probabilistic_2023}.

These models are continuous and generative. Three-dimensional
points along with intensity values can be sampled from the model
using:
\begin{minted}[bgcolor=bg,breaklines,fontsize=\footnotesize]{python}
# Sample 640*480 points from the model
rp = sg_gpu.joint_dist_sample(640*480)
\end{minted}
A plot of the resampled point cloud is shown in~\cref{sfig:resampled}.

If the 3D point locations are known, the expected intensity values and variance
can be inferred from the model at these locations:
\begin{minted}[bgcolor=bg,breaklines,fontsize=\footnotesize]{python}
# locs is a N x 3 numpy array
# E is N x 1 expected intensities
# V is N x 1 variance
_, E, V = mgpu.color_conditional(locs)
\end{minted}
A plot of intensity values \texttt{E} is shown in~\cref{sfig:inferred}.

The SOGMM model is compact and its size can be computed as follows.
\begin{minted}[bgcolor=bg,breaklines,fontsize=\footnotesize]{python}
# computing memory usage
M = mgpu.n_components_

# 4 bytes per float
# 1 float value per weight
# 4 float values per mean
# 10 float values per covariance
mem_bytes = 4 * M * (1 + 10 + 4)
\end{minted}
which is $69.78$ kilobytes for the model learnt in~\cref{fig:gira3d-recon}.

\begin{figure*}
      \centering
      \footnotesize
      \subfloat[\label{sfig:target-platforms}Target Platforms]{\input{figures/target_platforms_table.tex}}\\
      \ifthenelse{\equal{\arxivmode}{true}}
      {
      \subfloat[\label{sfig:rtx3090-sogmm}Ryzen/RTX3090, SOGMM]{\includegraphics[]{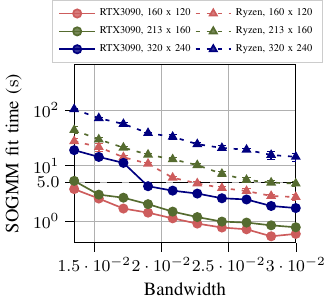}}%
      \subfloat[\label{sfig:rtx3060-sogmm}Intel/RTX3060, SOGMM]{\includegraphics[]{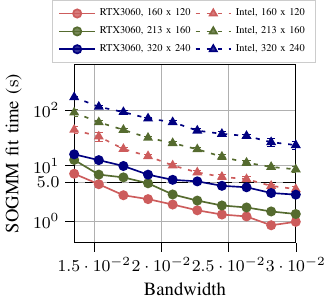}}%
      \subfloat[\label{sfig:orin-sogmm}ARM-12c/Orin, SOGMM]{\includegraphics[]{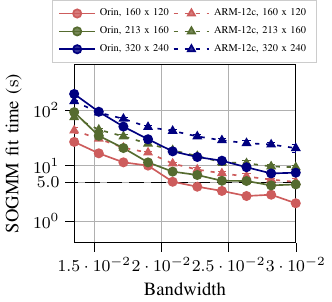}}\\
      \subfloat[\label{sfig:xavier-sogmm}ARM-8c/Xavier, SOGMM]{\includegraphics[]{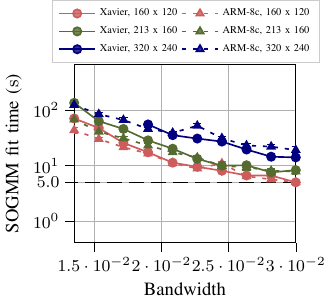}}
      \subfloat[\label{sfig:tx2-sogmm}ARM-6c/TX2, SOGMM]{\includegraphics[]{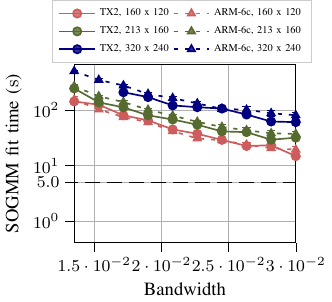}}
      \subfloat[\label{sfig:ryzen-scikit}Comparison with \texttt{scikit-learn} on Ryzen]{\includegraphics[]{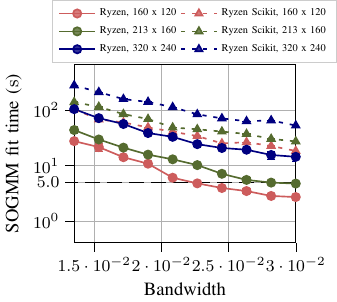}}
      }
      {
      \subfloat[\label{sfig:rtx3090-sogmm}Ryzen/RTX3090, SOGMM]{\input{figures/rtx3090-sogmm.tex}}%
      \subfloat[\label{sfig:rtx3060-sogmm}Intel/RTX3060, SOGMM]{\input{figures/rtx3060-sogmm.tex}}%
      \subfloat[\label{sfig:orin-sogmm}ARM-12c/Orin, SOGMM]{\input{figures/orin-sogmm.tex}}\\
      \subfloat[\label{sfig:xavier-sogmm}ARM-8c/Xavier, SOGMM]{\input{figures/xavier-sogmm.tex}}
      \subfloat[\label{sfig:tx2-sogmm}ARM-6c/TX2, SOGMM]{\input{figures/tx2-sogmm.tex}}
      \subfloat[\label{sfig:ryzen-scikit}Comparison with \texttt{scikit-learn} on Ryzen]{\input{figures/ryzen-scikit.tex}}
      }
      \caption{\label{fig:gira3d-recon-times}Comparison of SOGMM computation time via GIRA Reconstruction on the target platforms listed
            in~\cref{sfig:target-platforms}. In~\protect\subref{sfig:rtx3090-sogmm}
            and~\protect\subref{sfig:rtx3060-sogmm} the GPU-accelerated case
            on the desktop platforms provides more than an order of magnitude improvement
            in timing compared to the CPU-only case for most image sizes.
            The results of the embedded platforms shown in \protect\subref{sfig:orin-sogmm},~\protect\subref{sfig:xavier-sogmm}
            and~\protect\subref{sfig:tx2-sogmm} demonstrate that the relative performance improvements seem to
            degrade with increasing SWaP constraints.
            In any case,~\protect\subref{sfig:ryzen-scikit} shows that our CPU implementation performs
            nearly an order of magnitude faster than a reference SOGMM implementation
            using \texttt{scikit-learn}.}
\end{figure*}

The time taken to learn a SOGMM is reported as a function
of bandwidth parameter for a diverse set of platforms outlined in~\cref{sfig:target-platforms}.
The input data for this experiment corresponds
to frame 854, which was randomly selected, of the
simulated \texttt{livingroom1} data from the Augmented ICL-NUIM
datasets~\citep{choi_robust_2015}. Ten equally spaced bandwidth values
from $0.0135$ to $0.0300$ are used.  Image sizes of $320 \times 240$,
$213 \times 160$, and $160 \times 120$ are used (corresponding to $2
\times$, $3 \times$, and $4 \times$ reduction along each axis of the
original $640 \times 480$ image). Because there is randomness
in the KInit step, each case is run ten times and averaged to obtain
accurate timing results.

\Cref{sfig:rtx3090-sogmm,sfig:rtx3060-sogmm,sfig:orin-sogmm,sfig:xavier-sogmm,sfig:tx2-sogmm}
plot the results of these trials for the CPU-only (dashed
lines with triangle markers) and GPU-accelerated (solid lines with
circle markers) implementations. The y-axes of these plots
use a base-10 log-scale and the observed standard deviation,
which is plotted as error bars, is very low compared to the mean
values. From~\cref{sfig:rtx3090-sogmm,sfig:rtx3060-sogmm} we observe
over an order of magnitude faster performance when using the
GPU-accelerated version of the system for all image sizes. Further,
there is an overall decrease in performance from Ryzen/RTX3090
(\cref{sfig:rtx3090-sogmm}) to Intel/RTX3060
(\cref{sfig:rtx3060-sogmm}) for both CPU-only and GPU-accelerated
versions. This is expected due to the decrease in computational
capability for both the CPU and GPU.

\Cref{sfig:orin-sogmm} provides results for
ARM-12c/Orin platform. In this case, the gains for the GPU-accelerated version
are lower than the desktop platforms. Notice that for image size $320 \times
240$ the CPU starts performing better than GPU at low bandwidths. At low
bandwidths, the number of estimated components are high.

\Cref{sfig:xavier-sogmm,sfig:tx2-sogmm} suggest a further decrease in relative
performance improvement in using the GPU-accelerated version as opposed to the CPU-only
version of our system. Further, due to memory constraints the $320 \times 240$
image size fails for both platforms below certain bandwidths. Both ARM-8c/Xavier
and ARM-6c/TX2 are SWaP-constrained platforms used on robots.
For real-world usage of our framework, we recommend using the CPU-only
version when CPU resources are not required by other subsystems (e.g., planning,
control, and visual-inertial odometry) and using the GPU-accelerated version
when CPU resources are in demand (which is often the case).

%% file: figures/target_platforms_table.tex
\begin{tabular}{c|cccc}
  \textbf{Platform ID} & \textbf{Platform Type} & \textbf{CPU}                             & \textbf{GPU}            & \textbf{Memory (CPU/GPU)}                \\
  \hline
  Ryzen/RTX3090        & Desktop                & AMD Ryzen Threadripper 3960X, 48 threads & NVIDIA GeForce RTX 3090 & $\SI{252}{\giga\byte}$ / $\SI{24}{\giga\byte}$ \\
  Intel/RTX3060        & Desktop                & Intel Core i9-10900K, 20 threads         & NVIDIA GeForce RTX 3060 & $\SI{32}{\giga\byte}$ / $\SI{12}{\giga\byte}$  \\
  ARM-12c/Orin         & Embedded               & ARMv8 Processor rev 1 (v8l), 12 threads  & NVIDIA Jetson Orin      & $\SI{32}{\giga\byte}$                      \\
  ARM-8c/Xavier        & Embedded               & ARMv8 Processor rev 0 (v8l), 8 threads   & NVIDIA Jetson Xavier    & $\SI{16}{\giga\byte}$                      \\
  ARM-6c/TX2           & Embedded               & ARM Cortex-A57, 6 threads                & NVIDIA Jetson TX2       & $\SI{8}{\giga\byte}$
\end{tabular}

%% file: content/gira3d_registration.tex
\begin{figure}
  \centering
  \subfloat[Before registration\label{sfig:misaligned}]{\includegraphics[height=2.5cm,trim=0 0 0 0,clip]{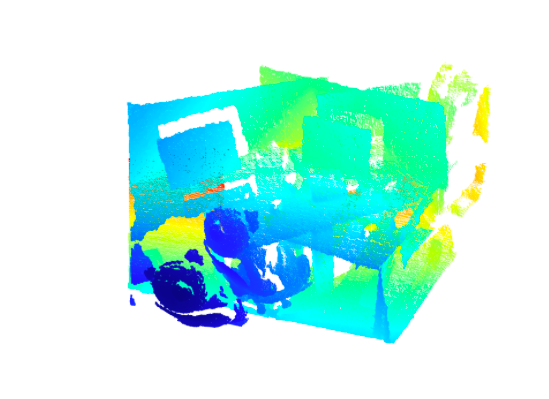}}\hfill%
  \subfloat[After registration\label{sfig:aligned}]{\includegraphics[height=2.5cm,trim=0 0 0 0,clip]{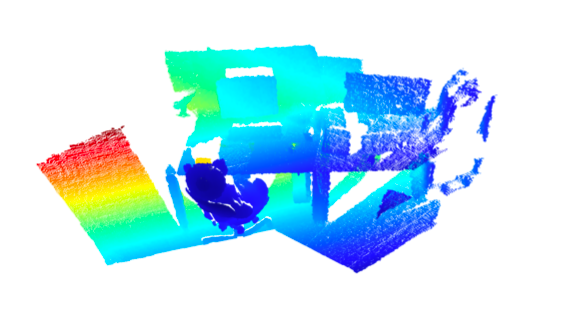}}
  \caption{\label{fig:single-pair-registration} The point clouds
    in~\protect\subref{sfig:misaligned} are originally
    misaligned.~\protect\subref{sfig:aligned} The code
    in~\cref{ssec:gira3d-registration} estimates the SE(3) transform to
    align them.}
  \vspace{-0.5cm}
\end{figure}
\subsection{GIRA Registration\label{ssec:gira3d-registration}} This
module implements (1) registering a pair of
GMMs~\cite{tabib_-manifold_2018} and (2) closing the loop using a pose
graph optimization~\cite{tabib_simultaneous_2021}.

\begin{figure}
  \centering
  \subfloat[Without loop closure\label{sfig:no-loop-closure}]{\includegraphics[height=4cm]{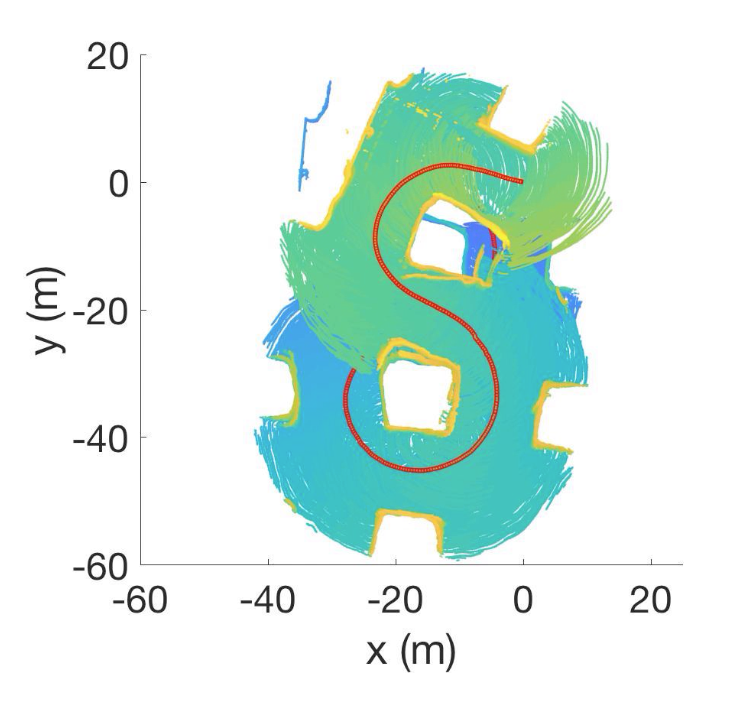}}%
  \subfloat[Loop closure\label{sfig:loop-closure}]{\includegraphics[height=4cm]{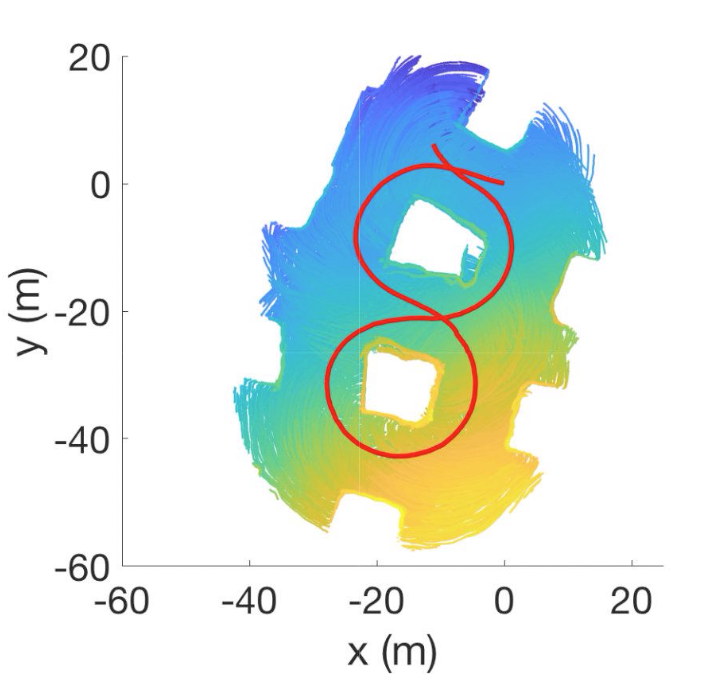}}
  \caption{\label{fig:pose-graph} The trajectories reconstructed
    using~\protect\subref{sfig:no-loop-closure} frame-to-frame
    registration and~\protect\subref{sfig:loop-closure} with loop closure
    is enabled are shown with the pointclouds plotted.}
\end{figure}

The \emph{anisotropic}, \emph{isoplanar}, and \emph{isoplanar-hybrid}
registration variants from~\cite{tabib_-manifold_2018} are implemented
in this module. Python and MATLAB interfaces have been developed,
  but this document provide examples only for the Python interface. The
isoplanar-hybrid registration approach first calls a coarse
optimizing using the isoplanar registration function followed by a
refinement optimization using the anisotropic registration. The source
and target variables are paths to files containing GMMs.
\begin{minted}[bgcolor=bg,breaklines,breakindentnchars=4,fontsize=\footnotesize,breaksymbolleft=]{python}
  from gmm_d2d_registration_py import isoplanar_registration
  from gmm_d2d_registration_py import anisotropic_registration

  # Initial registration guess
  Tinit = np.eye(4)

  # Isoplanar registration
  Tiso = isoplanar_registration(Tinit, source, target)
  Tout = anisotropic_registration(Tiso, source, target)

  # Rotation and translation solutions
  R = Tout[0:3, 0:3]
  t = Tout[0:3, 3]
\end{minted}
The result of registering a single pair of images may be seen
in~\cref{fig:single-pair-registration}. In addition, a pose graph
optimization example is provided, which uses GTSAM
\cite{dellaert2012gtsam}. A comparison of the frame-to-frame
registration with and without loop closure is shown
in~\cref{fig:pose-graph}.

%% file: content/gira3d_occupancy_modeling.tex
\begin{figure}
  \centering
  \includegraphics[width=0.7\linewidth,trim=70 40 70 120,clip]{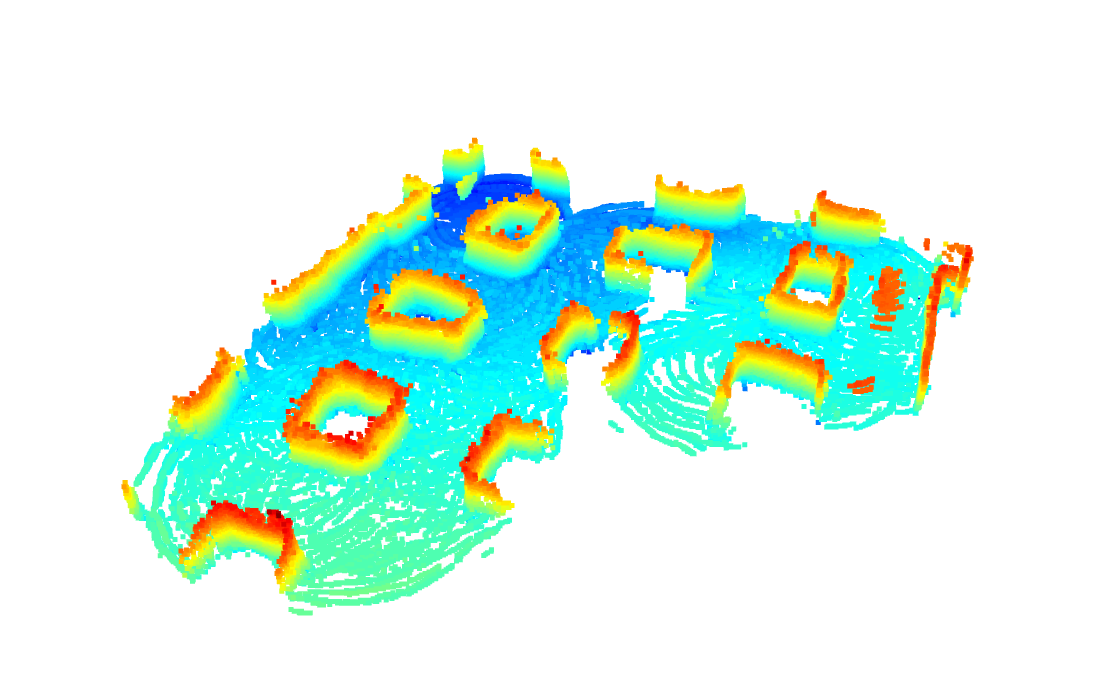}
  \caption{\label{fig:occupancy-modeling} Resampled points from
a GMM are added to an occupancy grid map and the occupied
voxels are queried and visualized.}
\vspace{-0.5cm}
\end{figure}
This module implements occupancy reconstruction by sampling from a GMM
and raytracing through an occupancy grid map. MATLAB and Python
interfaces are provided, but only the Python interface is discussed in
this document\footnote{Detailed documentation for both MATLAB and
Python is provided at \url{https://gira3d.github.io/docs/index.html}.}. Like the registration module
detailed in~\cref{ssec:gira3d-registration}, this module is compatible
with scikit-learn~\cite{scikit-learn} GMMs and assumes GMMs are loaded
from file.

\begin{minted}[bgcolor=bg,breaklines,breakindentnchars=4,fontsize=\footnotesize,breaksymbolleft=]{python}
  # Create 3d occupancy grid with parameters p
  grid = Grid3D(p)

  # Nx3 sampled from GMM (assumed in world frame)
  pts = gmm.sample(num_pts)

  # Add the points to the grid
  for i in range(0, num_pts):
    ray_end = Point(pts[i,0], pts[i,1], pts[i,2])

    # sensor_pose is in world frame
    # TRIMMED_MAX_RANGE set by user
    grid.add_ray(sensor_pose, ray_end, TRIMMED_MAX_RANGE)
\end{minted}

Functions for querying occupied, free, and unknown voxels are provided
through Python and MATLAB bindings of C++ code. The result of adding
sampled points from the Mine dataset GMMs and querying the occupied
voxels is shown in~\cref{fig:occupancy-modeling}.

%% file: content/impl_detail.tex
\begin{figure}
  \ifthenelse{\equal{\arxivmode}{true}}
  {
  \includegraphics[width=\linewidth]{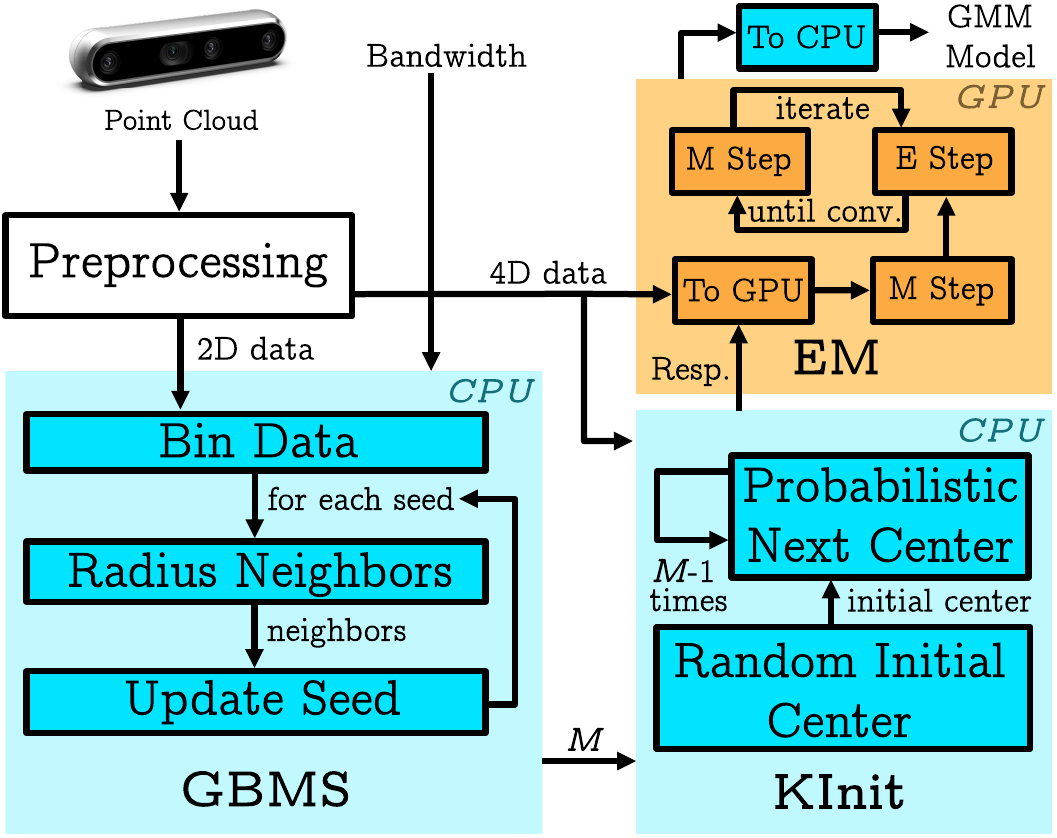}%
  }
  {
  \includegraphics[width=\linewidth]{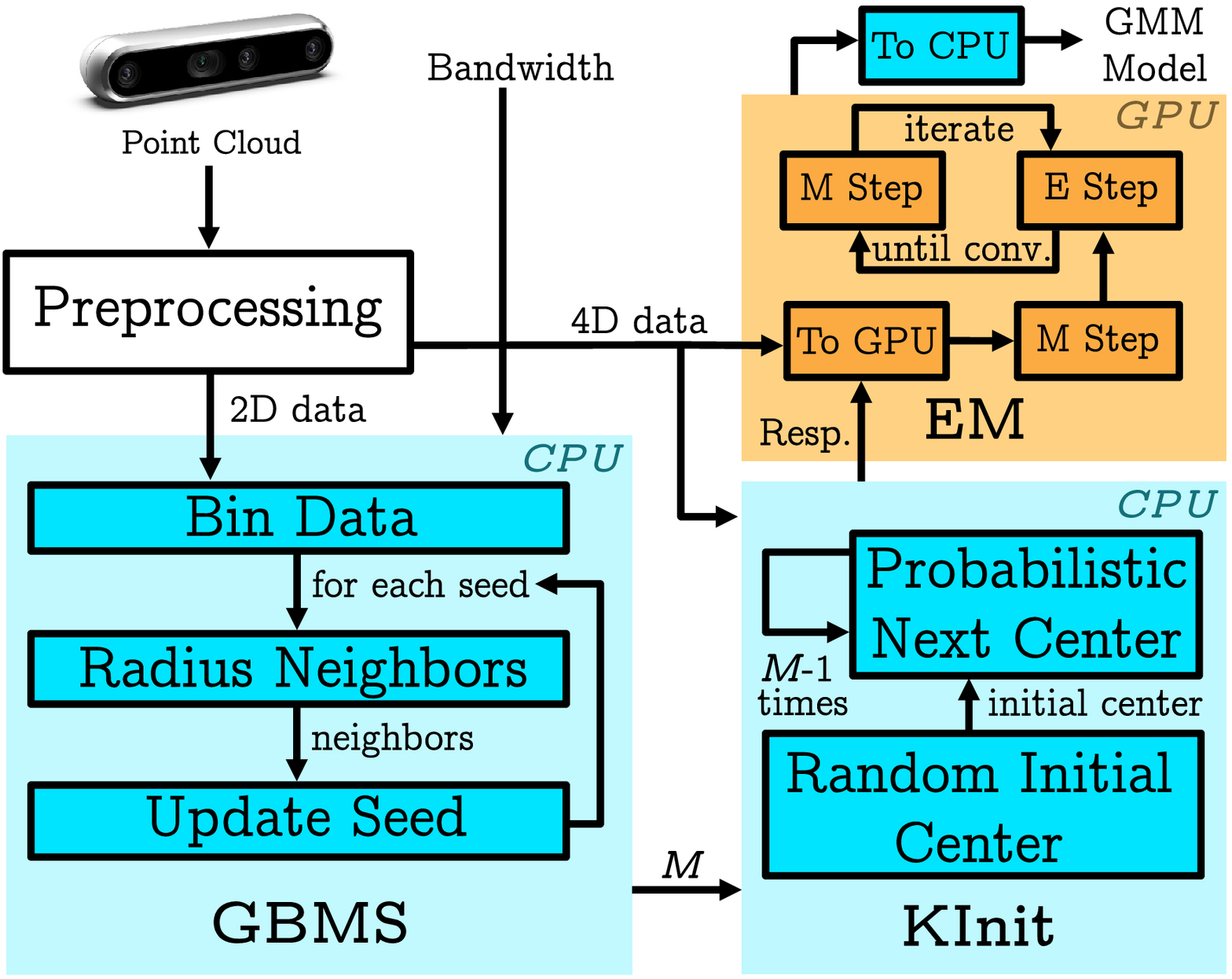}%
  }
  \caption{\label{fig:information-flow}Information flow for the
    GPU-accelerated adaptive point cloud modeling system. Given a
      bandwidth parameter and depth-intensity image pair, the Gaussian
    Blurring Mean Shift (GBMS) obtains the number of components
    $\nComponents$.  The number of components and the 4D data are used by
    KInit to calculate the responsibility matrix used by the EM
    algorithm. The result of the EM algorithm is the SOGMM
    model~\citep{goel_probabilistic_2023}.}
\vspace{-0.5cm}
\end{figure}

This section details the GPU-accelerated software architecture
of GIRA Reconstruction.~\Cref{fig:information-flow} provides an
overview of the accelerated SOGMM
components~\citep{goel_probabilistic_2023}.

\paragraph{Gaussian Blurring Mean Shift}
\citet{comaniciu_mean_2002} leverage a binned estimator to
determine seeds for the algorithm. A
kd-tree~\cite{blanco2014nanoflann} is used to query the neighbors in
$\datasetY$. The points within the specified radius are averaged and
the seed is updated to the new location. The algorithm terminates
when either the number of maximum iterations is reached or
there is no substantial change with respect to the previous seed
position.

\paragraph{Expectation Maximization} The EM algorithm consists of the
Expectation (E) and Maximization (M) steps. The E Step
evaluates the responsibilities $\resp_{nb}$ using the current
parameters $\mean_b$, $\cov_b$ and $\weight_b$ via
\begin{align}
  \resp_{nb} &= \frac{\weight_b \gaussian{\pointx_n}{\mean_b}{\cov_b}{}}
               {\sum\limits_{a=1}^{\nComponents} \weight_a \gaussian{\pointx_n}{\mean_a}{\cov_a}{}}.
               \label{eq:estep}
\end{align}
To reduce the computational complexity of~\cref{eq:estep}, the natural logarithm
can be applied to convert the multiplications and divisions into sums and
differences:
\begin{align}
  \ln \resp_{nb} &= \ln \weight_b + \ln \left( \gaussian{\pointx_n}{\mean_b}{\cov_b}{} \right) \notag \\
                 &~~~~~~~ - \ln \left( \sum\limits_{a=1}^{\nComponents} \weight_a \gaussian{\pointx_n}{\mean_a}{\cov_a}{} \right).
  \label{eq:logestep}
\end{align}
Term 2 of~\cref{eq:logestep} may be rewritten, as derived
in~\cite{buitinck2013api,pedregosa2011scikit,wenniethesis},
\begin{footnotesize}
\begin{align}
  &\ln \left( \gaussian{\pointx_n}{\mean_b}{\cov_b}{} \right) \notag\\
  &=-\frac{1}{2}\left( D\ln(2\pi) + \sum\limits_{j=1}^D \left( \precchol_b(\pointx_n - \mean_b)  \right)_j^2 \right) + \left( \sum_{j=1}^D \ln \left(\text{diag}(\precchol_b)\right)_j \right)
  \label{eq:simplifiedGaussian}
\end{align}
\end{footnotesize}
where $\cov = \lowerL\lowerL^{\top}$, $\precchol = \lowerL^{-1}$, and $\lowerL$ is a
lower triangular matrix calculated using the Cholesky decomposition of the
covariance matrix.  Summing the logarithm of the diagonal entries of
$\precchol_b$ (i.e., $\ln(\text{diag}(\precchol_b))$)
in~\cref{eq:simplifiedGaussian} is equivalent to $\ln|\cov_b|^{-1/2}$.

The GPU implementation leverages higher-order tensor representations (rank-$3$
and rank-$4$ tensors)\footnote{For the exposition of the GPU-accelerated
components, tensor conventions from TensorFlow~\citep{abadi2016tensorflow} are
used.}. The weights are represented as a rank-$3$ tensor of shape $(1,
\nComponents, 1)$, means are represented as a rank-$3$ tensor of shape $(1,
\nComponents, 4)$, and covariances are represented as a rank-$4$ tensor of shape
$(1, \nComponents, 4, 4)$. This implementation accelerates unary (e.g.,
logarithm and exponential of a matrix, reduction operations like
summing along a dimension or taking a maximum along a dimension of a
rank-$2$ or a rank-$3$ tensor) and binary (e.g., addition,
subtraction, multiplication, and division of rank-$2$ and rank-$3$
tensors) operations via element-wise CUDA kernels with fixed block and
grid sizes for all GPUs.

Rank-$2$ tensor multiplication is accelerated using the
\texttt{cuBLAS}\footnote{\texttt{cuBLAS} \url{https://docs.nvidia.com/cuda/cublas}}
\texttt{gemm} routine. Rank-$3$ tensor multiplication is
accelerated via the \texttt{cuBLAS} \texttt{gemmStridedBatched} routine. The
Cholesky decomposition of a rank-$2$ tensor is accelerated via the
\texttt{cuSOLVER}\footnote{\texttt{cuSOLVER} \url{https://docs.nvidia.com/cuda/cusolver}}
\texttt{potrf} routine.  The Cholesky decomposition of a rank-$3$ tensor is
accelerated using the \texttt{cuSOLVER} \texttt{potrfBatched} routine.  Using
the Cholesky decomposition, a linear system of equations involving rank-$2$
tensors is solved using the \texttt{cuSOLVER} \texttt{potrs} routine and for rank-$3$ tensors using
the \texttt{potrsBatched} routine.








%% file: content/conclusion.tex
GIRA is a set of tools and software for processing point cloud data into
Gaussian mixture models for inference and robot autonomy. These tools and
software are released open-source \url{https://github.com/gira3d}. Fundamental robotics
capabilities \blue{from our prior works on point cloud
modeling~\citep{goel_probabilistic_2023}, pose
estimation~\citep{tabib_-manifold_2018,tabib_simultaneous_2021}, and occupancy
modeling~\citep{tabib_autonomous_2021} are included in the open-source release.}
These fundamental capabilities have applications beyond exploration and aerial
robotics. The adaptivity of the SOGMM representation has applicability to
perception in the small and fine grained manipulation tasks. The variable
resolution occupancy grid mapping and distribution to distribution registration
software may be leveraged for high-speed mobile robot applications like
\blue{off-road operations}. By releasing this software, the authors hope to
increase the accessibility of these formulations to technical experts.